\def\paperID{14196}
\def\confName{ICCV}
\def\confYear{2025}

\def\paperTitle{Early Timestep Zero-Shot Candidate Selection for Instruction-Guided Image Editing}

\def\authorBlock{
Joowon Kim$^{1}$\thanks{Equal contribution}
    \quad
    Ziseok Lee$^{2}$\footnotemark[1]
    \quad
    Donghyeon Cho$^{1}$
    \quad
    Sanghyun Jo$^{3}$
    \\
    {Yeonsung Jung$^{1}$
    \quad
    Kyungsu Kim$^{1,2}$\thanks{Corresponding Author}
    \quad
    Eunho Yang$^{1}$\footnotemark[2]} \\ \\
    $^{1}$ KAIST \quad
    $^{2}$ Seoul National University \quad 
    $^{3}$ OGQ \\ \\
    \small\texttt{kjwispro@kaist.ac.kr, ziseoklee@snu.ac.kr, hyeon9698@kaist.ac.kr, shjo.april@gmail.com} \\ \small\texttt{ys.jung@kaist.ac.kr, kyskim@snu.ac.kr, eunhoy@kaist.ac.kr}
}

\newif\ifreview 
\newif\ifarxiv \newcommand{\arxiv}{\arxivtrue}
\newif\ifcamera 
\newif\ifrebuttal 

\arxiv 

\pdfoutput=1
\documentclass[10pt,twocolumn,letterpaper]{article}
\ifreview \usepackage[review]{cvpr} \fi
\ifarxiv \usepackage[pagenumbers]{cvpr} \fi
\ifrebuttal \usepackage[rebuttal]{cvpr} \fi
\ifcamera \usepackage{cvpr} \fi

\usepackage{graphicx}	
\usepackage{amsmath}	
\usepackage{amssymb}	
\usepackage{booktabs}
\usepackage{times}
\usepackage{microtype}
\usepackage{epsfig}
\usepackage{caption}
\usepackage{float}
\usepackage{placeins}
\usepackage{color, colortbl}
\usepackage{stfloats}
\usepackage{enumitem}
\usepackage{tabularx}
\usepackage{xstring}
\usepackage{multirow}
\usepackage{xspace}
\usepackage{url}
\usepackage{subcaption}
\usepackage{xcolor}
\usepackage[hang,flushmargin]{footmisc}
\usepackage{pifont}
\newcommand{\cmark}{\ding{51}}%
\newcommand{\xmark}{\ding{55}}%
\usepackage{makecell}
\usepackage{kotex}
\usepackage{fvextra}
\usepackage[accsupp]{axessibility}

\ifcamera \usepackage[accsupp]{axessibility} \fi

% ...

\ifarxiv  \fi

\newcommand{\R}[1]{{%
    \textbf{%
        \ifstrequal{#1}{1}{\textcolor{red}{R#1}}{%
        \ifstrequal{#1}{2}{\textcolor{blue}{R#1}}{%
        \ifstrequal{#1}{3}{\textcolor{magenta}{R#1}}{%
        \ifstrequal{#1}{4}{\textcolor{teal}{R#1}}{%
                           \textcolor{cyan}{R#1}%
        }}}}%
    }%
}}

\newcommand\red[1]{{\color{red}#1}}
\newcommand\blue[1]{{\color{blue}#1}}

\usepackage{algorithm}
\usepackage{algpseudocode}  

\usepackage{xr-hyper}

\makeatletter
\newcommand*{\addFileDependency}[1]{
  \typeout{(#1)}
  \@addtofilelist{#1}
  \IfFileExists{#1}{}{\typeout{No file #1.}}
}

\makeatother
\newcommand*{\myexternaldocument}[1]{
    \externaldocument{#1}
    \addFileDependency{#1.tex}
    \addFileDependency{#1.aux}
}

\definecolor{cvprblue}{rgb}{0.21,0.49,0.74}
\usepackage[pagebackref,breaklinks,colorlinks,allcolors=cvprblue]{hyperref}
\usepackage[capitalize]{cleveref}
\crefname{section}{Sec.}{Secs.}
\crefname{table}{Table}{Tables}
\crefname{figure}{Fig.}{Figs.}

\ifarxiv \crefname{appendix}{App.}{Apps.}
\else \crefname{appendix}{Suppl.}{Suppls.} \fi

\unless\ifarxiv \myexternaldocument{_supplementary} \fi

\begin{document}
\title{\paperTitle}
\author{\authorBlock}

\makeatletter
\let\@oldmaketitle\@maketitle
    \renewcommand{\@maketitle}{\@oldmaketitle
    \vspace*{-20pt}
    \centering
    \includegraphics[width=\textwidth]{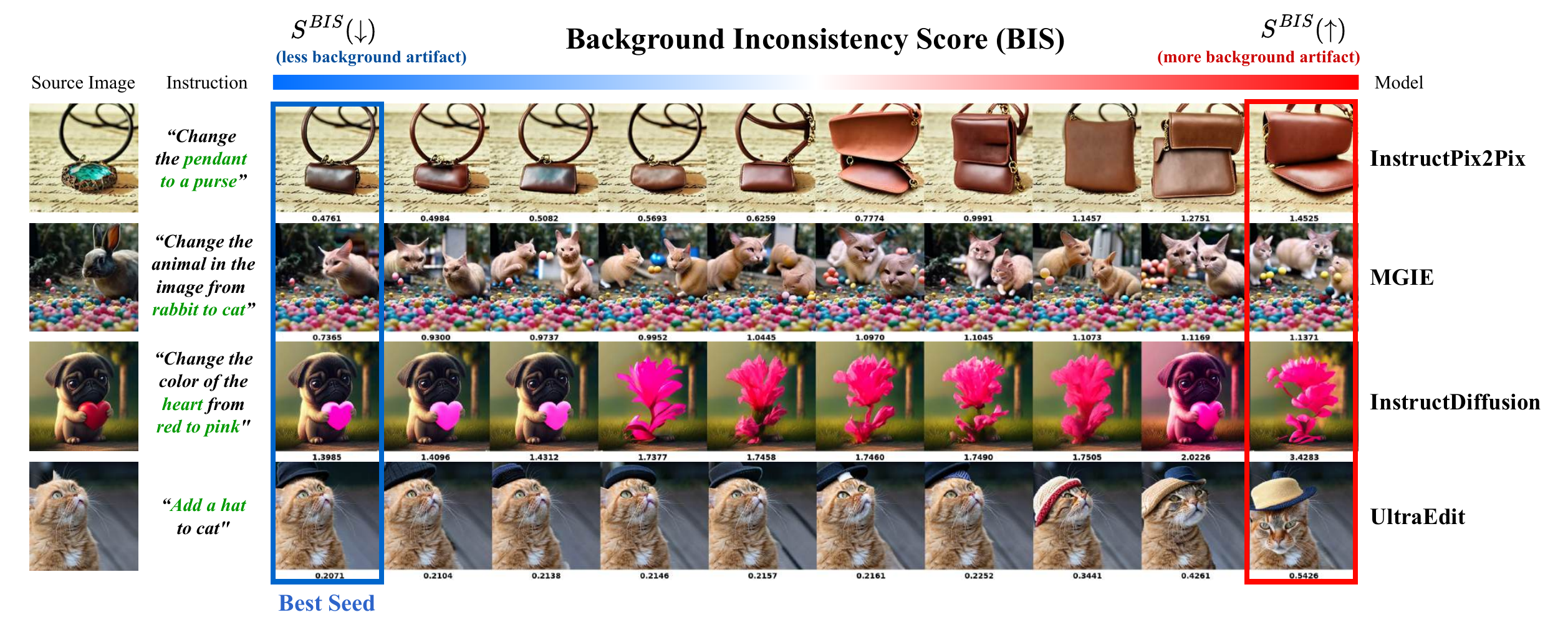}
    \vspace{-0.8cm}
    \captionof{figure}{Instruction-guided image editing models are highly influenced by the noise seed. To address this issue, we propose a unique candidate selection method (ELECT), which is successfully selects the best seed for background consistency while maintaining the editability of the base model. We propose Background Inconsistency Score ($S^\text{BIS}$) that quantifies the degree of unintended background changes in an edited image, measuring relatively how well the background is preserved compared to other candidates in a self-supervised manner. Samples with low $S^\text{BIS}$ (\blue{blue}) have a consistent background while samples with high $S^\text{BIS}$ (\red{red}) display multiple artifacts and distortion.}
    \label{fig:seed_variance}
  \bigskip}
\makeatother
\maketitle

\begin{abstract}

Despite recent advances in diffusion models, achieving reliable image generation and editing results remains challenging due to the inherent diversity induced by stochastic noise in the sampling process. Particularly, instruction-guided image editing with diffusion models offers user-friendly editing capabilities, yet editing failures, such as background distortion, frequently occur across different attempts. Users often resort to trial and error, adjusting seeds or prompts to achieve satisfactory results, which is inefficient.
While seed selection methods exist for Text-to-Image (T2I) generation, they depend on external verifiers, limiting their applicability, and evaluating multiple seeds increases computational complexity, reducing practicality.
To address this, we first establish a new multiple-seed-based image editing baseline using background consistency scores, achieving Best-of-N performance without supervision. Building on this, we introduce \textbf{ELECT} (\textbf{E}arly-timestep \textbf{L}atent \textbf{E}valuation for \textbf{C}andidate selec\textbf{T}ion), a zero-shot framework that selects reliable seeds by estimating background mismatches at early diffusion timesteps, identfying the seed that retains the background while modifying only the foreground. ELECT ranks seed candidates by a background inconsistency score, filtering unsuitable samples early based on background consistency while fully preserving editability. Beyond standalone seed selection, ELECT integrates into instruction-guided editing pipelines and extends to Multimodal Large-Language Models (MLLMs) for joint seed + prompt selection, further improving results when seed selection alone is insufficient. Experiments show that ELECT reduces computational costs (by 41\% on average and up to 61\%) while improving background consistency and instruction adherence, achieving around 40\% success rates in previously failed cases—without any external supervision or training. Our code is available at \url{https://github.com/Joow0n-Kim/ELECT}

\end{abstract}

\vspace{-1em}
\section{Introduction}
\label{sec:intro}

Instruction-guided image editing \cite{IP2P_brooks2023instructpix2pix,MagicBrush_NEURIPS2023_64008fa3,InstructDiff_Geng23instructdiff,EmuEdit_sheynin2024emu,DialogPaint_wei2023dialogpaint,HIVE_zhang2024hive,MGIE_fu2024guiding,SmartEdit_Huang_2024_CVPR,HQ-Edit_hui2024hqedithighqualitydatasetinstructionbased,SEED-Data-Edit_ge2024seeddataedittechnicalreporthybrid,EditWorld_yang2024editworldsimulatingworlddynamics,MoEController_li2024moecontrollerinstructionbasedarbitraryimage,UltraEdit_zhao2024ultraedit} enables fine-grained modifications based on textual prompts, with applications in content creation and design. However, diffusion-based editing remains unreliable due to the inherent stochasticity of text-to-image models, which produce varying outputs depending on the initial random noise, leading to unpredictable outcomes \cite{DiffRS_na2024diffusion,InitNO_guo2024initno,GIS_Mao_2023,InitIC_mao2023semantic,Good_Seed_xu2024good,FIND_chen2024find}. Consequently, users must manually sift through multiple generations to find a suitable output, which makes the editing process inefficient and results in inconsistent modifications.

Auto-regressive models like LLMs face the similar issue, sampling multiple outputs at inference to improve quality \cite{Snell2024ScalingLT}. Similarly, in text-to-image generation, search techniques such as \textit{Best of N}, which select the best result using specific verifiers, have been explored \cite{ma2025inference}. In instruction-guided image editing, users often generate multiple outputs by varying the random seed and manually selecting the most suitable result. However, this process is computationally expensive, with performance scaling linearly with N, making it impractical for real-time applications. Moreover, there is no established framework for efficiently identifying the optimal seed before full inference, underscoring the need for a more effective selection method.

Existing seed selection methods for text-to-image (T2I) generation, such as rejection sampling \cite{DiffRS_na2024diffusion}, seed optimization \cite{InitNO_guo2024initno,FIND_chen2024find}, and noise resampling \cite{GIS_Mao_2023,InitIC_mao2023semantic}, focus on image quality and prompt fidelity but overlook background consistency. These methods are unsuitable for instruction-guided editing, as they assess generation quality in isolation, without ensuring structural alignment with a reference image. Other studies \cite{Good_Seed_xu2024good, ma2025inference} still rely on external verifiers\textemdash such as Aesthetic scoring and CLIPScore \cite{hessel2021clipscore}\textemdash that require full inference, which makes them impractical for early-stage filtering.

To bridge this gap, we found that selecting the seed with the lowest background Mean Squared Error (MSE)\textemdash MSE computed over the masked background regions between the edited and source images\textemdash effectively reduces artifacts and improves instruction adherence—all without requiring additional models or supervision.  Since directly computing background MSE requires ground truth (GT) masks, which are unavailable at inference time, we leverage aggregated relevance maps as a proxy for GT masks, achieving performance parity with the GT-based approach across all metrics.

Despite its effectiveness, using relevance maps incurs high computational costs due to the evaluation of multiple seeds. To mitigate this, we analyzed the denoising process and observed that early timesteps already identify key regions for editing, with later steps refining the details. These insights enable an \textbf{early-timestep evaluation strategy} that extracts a background mask and estimates the final output using Tweedie’s formula. This early evaluation strategy significantly reduces computational cost while maintaining or even surpassing the performance of full inference-based selection.

Hence, we propose \textbf{E}arly-timestep \textbf{L}atent \textbf{E}valuation for \textbf{C}andidate Selec\textbf{T}ion (\textbf{ELECT}), a zero-shot framework for selecting optimal seeds in image-to-image (I2I) editing. Unlike text-to-image (T2I) generation, where external verifiers are often needed to assess image quality post-generation, I2I editing is conditioned on a source image, allowing us to evaluate seed suitability directly from early-timestep diffusion latents. We also propose Background Inconsistency Score (BIS) as a lightweight selection metric that measures unwanted background changes. ELECT estimates BIS from early timestep latents and selects the optimal candidate, significantly reducing computational cost. Unlike prior T2I methods, ELECT requires no external models, additional training, or full inference, making it lightweight, model-agnostic, and easily integrable into existing pipelines. Furthermore, we extend ELECT beyond seed selection to prompt selection by incorporating multimodal large language models (MLLMs) \cite{GPT4V_yang2023dawn, gpt4o}, enhancing editing reliability when seed selection alone is insufficient.

Our contributions are summarized as follows:
\begin{itemize}
    \item We introduce Background Inconsistency Score (BIS), a metric for measuring unwanted background changes without requiring external verifiers or full inference.

    \item We propose ELECT, the first inference-time candidate selection framework for instruction-guided image editing, enabling efficient seed selection by quantifying background inconsistency at early denoising steps and reducing computational costs by 41\% on average and up to 61\% NFE while maintaining or surpassing full inference performance.
    
    \item We extend ELECT to joint seed and prompt selection using multimodal large language models (MLLMs) \cite{gpt4o} refining out-of-distribution instructions and improving VIEScore by $+0.56$ on average. 
\end{itemize}

\section{Related Work}
\label{sec:related}

Early works in text-guided image editing leveraged source-target caption pairs \cite{P2P_hertz2022prompt, PnP_Inversion_ju2023direct, MasaCtrl_Cao_2023_ICCV, InfEdit_xu2023infedit, PostEdit_tian2024postedit, Eta_Inversion_kang2025eta, DDPM_Inversion_huberman2024edit, TODInv_xu2024task, SPDInv_li2025source} with attention modulation techniques. Recently, instruction-guided editing methods have gained attention as it replaces source and target captions with a single instruction prompt and eliminates hyperparameters involved in attention modulation. Our work focuses on further improving the user-friendliness of instruction-guided editing, particularly addressing the challenge users face when selecting the appropriate seed and prompt for optimal background inconsistency.

\textbf{Instruction-Guided Image Editing with Diffusion Models.} Unlike caption-based approaches, instruction-guided editing methods \cite{IP2P_brooks2023instructpix2pix, MagicBrush_NEURIPS2023_64008fa3, InstructDiff_Geng23instructdiff, EmuEdit_sheynin2024emu, DialogPaint_wei2023dialogpaint, HIVE_zhang2024hive, MGIE_fu2024guiding, SmartEdit_Huang_2024_CVPR, HQ-Edit_hui2024hqedithighqualitydatasetinstructionbased, SEED-Data-Edit_ge2024seeddataedittechnicalreporthybrid, EditWorld_yang2024editworldsimulatingworlddynamics, MoEController_li2024moecontrollerinstructionbasedarbitraryimage, UltraEdit_zhao2024ultraedit} take an input image $I$ and a textual command $T$ (e.g., "add a dog") to guide modifications. InstructPix2Pix (IP2P) \cite{IP2P_brooks2023instructpix2pix} uses GPT-3 \cite{GPT3_brown2020language} and Prompt2Prompt \cite{Prompt2Prompt_hertz2022prompt} to create a dataset, and trains a denoising network conditioned on edit instructions and the original image. Subsequent studies such as MagicBrush \cite{MagicBrush_NEURIPS2023_64008fa3}, UltraEdit \cite{UltraEdit_zhao2024ultraedit}, HIVE \cite{HIVE_zhang2024hive}, and HQ-Edit \cite{HQ-Edit_hui2024hqedithighqualitydatasetinstructionbased} propose fine-tuning techniques for IP2P through improved datasets, verifier models \cite{GPT4V_yang2023dawn}, and RLHF \cite{RLHF_li2023reinforcement}. InstructDiffusion (InsDiff) \cite{InstructDiff_Geng23instructdiff} proposes a unified framework across multiple computer vision tasks including segmentation and editing, and MGIE \cite{MGIE_fu2024guiding} and SmartEdit \cite{SmartEdit_Huang_2024_CVPR} leverages multi-modal LLMs (MLLMs) to guide and enhance image editing. Although instruction-guided methods outperform previous methods, they tend to overedit images and introduce variability due to sensitivity to initial seeds and instruction phrasing, leading to inconsistent edits across runs. To address this, Watch Your Steps \cite{Watch_Your_Steps_mirzaei2025watch}, Focus on Your Instruction \cite{FoI_guo2024focus}, and ZONE \cite{ZONE_li2024zone} employ mask-guided approaches that restrict modifications to a thresholded foreground mask. However, these methods rely on a fixed mask obtained from a single seed, failing to account for the inherent seed variability of diffusion models. Consequently, errors in the mask can lead to significant background inconsistencies.

\textbf{Candidate Selection for Diffusion Models.}
Best-of-N is a well-established alignment strategy in inference-time scaling for LLMs \cite{BoN_sun2025fast}. Recent advances have extended inference-time scaling to diffusion models \cite{ma2025inference, CoT_guo2025can}, demonstrating the effectiveness of reward models that evaluate the final generated outputs and select the best-of-N result to enhance generation quality. Prior work in T2I generation has explored candidate seed selection \cite{GIS_Mao_2023, InitIC_mao2023semantic, Good_Seed_xu2024good, ma2025inference} and optimization techniques \cite{InitNO_guo2024initno, FIND_chen2024find}, demonstrating that seed choice significantly impacts output quality. However, existing approaches only apply to T2I generation tasks and rely on computationally expensive external models as verifiers \cite{GPT4V_yang2023dawn}. We bridge this gap by introducing a best-of-N selection framework for selecting optimal seeds for I2I editing task.

\vspace{-0.3em}
\section{Preliminaries}
\label{sec:preliminaries}

\textbf{Diffusion Models.} Diffusion models \cite{song2021scorebased} involve two processes: a forward process adding noise and a reverse denoising process. Discretized into $T$ timesteps, noise $z_t$ is generated with coefficients $\bar{\alpha}_t$:
\begin{equation}
\label{eq:addnoise}
    z_t = \sqrt{\bar{\alpha}_t}z_0 + \sqrt{1-\bar{\alpha}_t} \epsilon
\end{equation}
where $\epsilon \sim \mathcal{N}(0,I), t=1,\dots,T$. A neural network, $\epsilon_\theta(z, t)$, estimates the noise $\epsilon$ for reverse denoising, producing a denoised image $\hat{z}_0$ using a reverse transformation (i.e., Tweedie's formula):
\begin{equation}
\label{eq:reconstruct}
    \hat{z_0} = \frac{1}{\sqrt{\bar{\alpha}_{t}}} (z_t - \sqrt{1-\bar{\alpha}_t} \epsilon_{\theta^\star}(z_t, t))
\end{equation}

\textbf{InstructPix2Pix.} Recent text-to-image diffusion models \cite{SD_rombach2022high, SD3_esser2024scaling} are trained on text-conditioned datasets, enabling conditional generation $\epsilon_\theta(z_t,t,C_T)$ on text embedding $C_T$. InstructPix2Pix (IP2P) \cite{IP2P_brooks2023instructpix2pix}, is a text-conditioned diffusion model fine-tuned on an instruction-based dataset. Built on latent diffusion \cite{SD_rombach2022high}, IP2P learns to modify images by conditioning on both the original image $I$ and an edit instruction $C_T$, enabling image-conditional generation $\epsilon_\theta (z_t,t,I,C_T)$. The strength of the edit can be controlled by the image guidance scale, $s_I$ and the text guidance scale $s_T$. The final score estimate is then obtained as 
\begin{equation}
\begin{aligned}
\tilde{\epsilon}_\theta(z_t,t,I,C_T)=& \epsilon_\theta(z_t,t,\emptyset_I,\emptyset_T)\\
    + s_I(\epsilon_\theta&(z_t,t,I,\emptyset_T)-\epsilon_\theta(z_t,t,\emptyset_I,\emptyset_T) )\\
    +s_T( \epsilon_\theta&(z_t,t,I,C_T)-\epsilon_\theta(z_t,t,I,\emptyset_T)  )
\end{aligned}
\end{equation}

\textbf{Edit Relevance Map.}
To evaluate the impact of an edit instruction on an image, we leverage an edit relevance map, first introduced by WYS \cite{Watch_Your_Steps_mirzaei2025watch}, which estimates the likelihood of each pixel being modified. This map serves as a crucial tool for identifying regions most affected by the edit process.

Given a source image $I$ and an edit instruction $C_T$, WYS constructs the relevance map by first adding noise to the encoded image representation $\mathcal{E}(I)$:
\begin{equation}
    z_{t}=\sqrt{\alpha_{t}}\mathcal{E}(I)+\sqrt{1-\alpha_{t}}\epsilon
\label{eq:add_noise_to_I}
\end{equation}
where $\epsilon\sim\mathcal{N}(0,I)$ is random noise and $\alpha_t$ controls the noise level. The InstructPix2Pix \cite{IP2P_brooks2023instructpix2pix} denoising network, $\epsilon_\theta$, then predicts noise estimates for both the conditioned and unconditioned cases, $\epsilon_\theta(z_{t},t,I,C_T)$ and $\epsilon_\theta(z_{t},t,I,\emptyset)$. The pixel-wise magnitude of their difference provides an estimate of edit relevance $M_{t}=|\epsilon_\theta(z_{t},t,I,C_T)-\epsilon_\theta(z_{t},t,I,\emptyset)|$.  

Outlier values in $M_t$ are clamped using an interquartile range filter and normlized into $[0,1]$.

The same principle applies to rectified flow models \cite{Rectified_Flow_liu2022flow}, which replace diffusion with a velocity field $v_\theta$ learned in data space. In this framework, the relevance map is computed as $M_t = |v_\theta(z_t,t,I,C_T) - v_\theta(z_t,t,I,\emptyset)|$, where $v_\theta$ guides $z_t$ toward the target image. This allows the method to generalize beyond diffusion-based models. For additional details on rectified flow, see \cref{app:rectified_flow}.

\begin{figure}[ht]
    \centering
    \includegraphics[width=\linewidth]{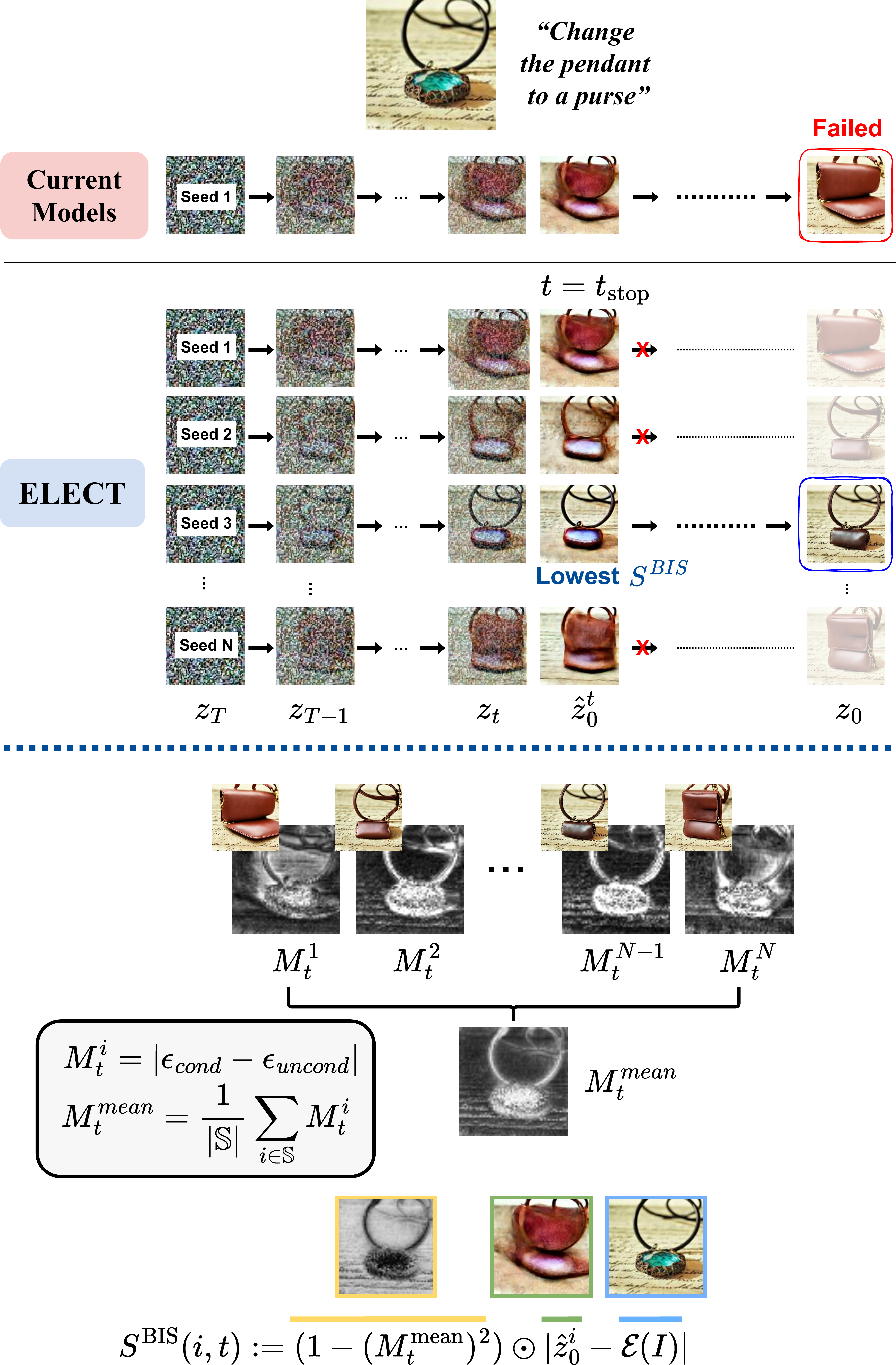}
    \caption{\textbf{Overview of the ELECT pipeline (\textit{top}) and details of Background Inconsistency Score (BIS) computation (\textit{bottom}).} The top panel illustrates candidate selection via early stopping and BIS evaluation. The bottom panel details BIS computation, including crowd-sourced reference masks and background masking, ensuring consistent edits with minimal distortions. The BIS metric compares clean images with the original input to quantify background distortions, ensuring consistent edits while minimal undesired changes.}
    \label{fig:overall_pipeline}
    \vspace{-0.4cm}
\end{figure}

Our approach modifies the original WYS formulation by eliminating the explicit noise perturbation in \eqref{eq:add_noise_to_I}. Instead, we extract $z_t$ directly from the intermediate denoised trajectory, reducing computational redundancy. Since deterministic samplers like DDIM and rectified flow rely only on the initial seed for stochasticity, we avoid unnecessary noise injections while maintaining reliable edit relevance estimation (see \cref{fig:overall_pipeline}).

\begin{figure}[tp]
    \centering
    \includegraphics[width=\linewidth]{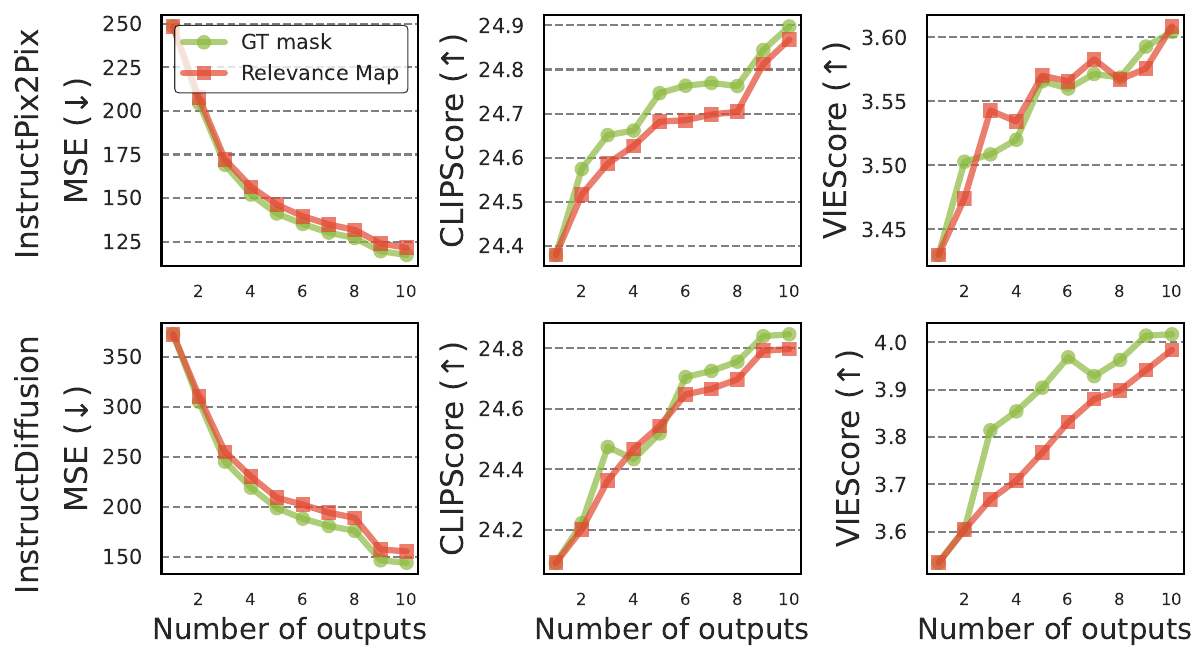}
    \vspace{-0.4cm}
    \caption{\textbf{Performance comparison of Best of N with GT mask (\textcolor{Green}{green}) vs. Best of N with Relevance Map (\red{red}).} Best of N chooses outputs with the lowest background inconsistency computed using either GT masks (w/ pixel-annotation) or the aggregated relevance map (w/o pixel-annotation) \eqref{eq:M_t,mean}. Selecting the best sample based on the relevance map (\red{\textbf{Ours: red lines}}) yields improvements comparable to selection based on ground truth mask (\textcolor{Green}{\textbf{green lines}}),with performance improvements observed across all metrics and perspectives as the number of outputs grows.}
    \label{fig:metric_w_mask}
    \vspace{-0.3cm}
\end{figure}

\section{Method}
\label{sec:method}
We introduce ELECT, a \textit{model-agnostic} and \textit{efficient} framework for selecting high-quality edited images from diffusion-based editing pipelines. ELECT stops denoising early and selects the best candidate based on background consistency.

\subsection{Observations}

Image editing models exhibit high output variance across different seeds, with over-editing and distortion levels varying significantly (see \cref{fig:seed_variance}). Simply using ground truth (GT) masks, we measured background MSE to identify samples with minimal distortion. We found that selecting the seed with the smallest background MSE effectively reduces unnecessary artifacts and improves instruction following, even without additional training or modulation. These observations are reflected in \cref{fig:metric_w_mask} where we observed consistent enhancement across multiple metrics.

But measuring background MSE requires GT masks, which are unavailable during inference time. We found that aggregating relevance maps from multiple seeds gives a mask \eqref{eq:M_t,mean} that effectively identifies foreground regions, allowing us to replace GT masks with our aggregated relevance maps. \cref{fig:metric_w_mask} confirms that selecting the best sample using $S^\text{BIS}$ achieves improvements comparable to selection with GT mask-based MSE, with performance improving as N grows.

Finally, analysis of the denoising process revealed that the model identifies regions of interest for editing during early timesteps ($t_\text{stop}=100 \to 80$), while later steps refine fine-grained details (see \cref{fig:mask_ablation}). Building on this insight, we extracted a background mask and estimated the edited image using Tweedie’s formula at an early timestep $t_\text{stop}$ to estimate the background MSE. This enabled early selection of the most consistent sample, significantly reducing the computational cost of evaluating multiple candidates.

\subsection{Background Inconsistency Score (BIS)}
A well-edited image should retain the background while modifying only the foreground according to the given instruction. To quantify this, we define the \textbf{Background Inconsistency Score (BIS)}, denoted as $S^{\text{BIS}}$, which measures unwanted background changes. This score is meaningful not as an absolute value but rather in the context of relative comparison with other candidates.

Given a source image $I$, a text instruction $T$ for editing, and a set of $N$ candidate seeds $\mathbb{S}=\{1,2,...,N\}$, we define the mean relevance map at timestep $t$ as:
\begin{equation}
M_{t}^\text{mean} = \frac{1}{|\mathbb{S}|}\sum_{i\in\mathbb{S}} M_{t}^{i}
\label{eq:M_t,mean}
\end{equation}
where $M_{t}^{i}$ represents the edit relevance map for the $i$-th seed at timestep $t$. Regions that are consistently edited across all seeds tend to have higher values in $M_{t}^{\text{mean}}$, since the relevance map gives higher values to pixels that are more likely to be modified. Instead of thresholding the mask as in previous work \cite{Watch_Your_Steps_mirzaei2025watch}, we square the mean relevance map $(M_{t}^{\text{mean}})^2$ to sharpen and emphasize relative importance within the mask, while preserving the smoothness of its values. Then we determine the seed that yields minimal change in the background regions by computing $S^\text{BIS}(i,t)$\footnote{The full expression for BIS is $S^\text{BIS}(i,t\mid \mathbb{S},\epsilon_\theta,I,C_T)$. \eqref{eq:S^BIS} is a simplified expression.}, the Background Inconsistency Score of seed $i$ at timestep $t$:
\begin{equation}
    S^\text{BIS}(i,t) := (1 - (M_{t}^{\text{mean}})^2) \odot |\hat{z}_{0}^i - \mathcal{E}(I)|
\label{eq:S^BIS}
\end{equation} 

Here $\odot$ denotes the Hadamard product and $\hat{z}_{0}^i$ is the predicted denoised edited latent for the $i$-th seed computed at timestep $t$ via Tweedie's formula:
\begin{equation}
    \hat{z}_{0}^i=\frac{z_{t}^i-\sqrt{1-\alpha_t}\epsilon_\theta(z_{t}^i,t,I,C_T)}{\sqrt{\alpha_t}}
\end{equation}
The corresponding formula for rectified flow models is $\hat{z}_{0}^i=z_{t}^i-v_\theta(z_{t}^i,t,I,C_T)\cdot t$. The optimal seed $i^*_t$ for background consistency at timestep $t$ is obtained by,
\begin{equation}
    i^*_t= \underset{i\in \mathbb{S}}{\text{argmin }} S^\text{BIS}(i,t)
\end{equation}
Softening the noise mask prevents the misclassification of poorly preserved background regions as foreground, a common issue with thresholding. By using continuous weights to emphasize over-edited background regions and reduce weight on edited foreground regions, this method avoids threshold dependency and ensures robust seed selection across diverse cases. \cref{fig:metric_w_mask} shows that $S^\text{BIS}$ achieves performance on par with GT-based selection.

\subsection{Early-timestep Latent Evaluation for Candidate selecTion (ELECT) Pipeline}

\begin{algorithm}[h]
\caption{$\texttt{ELECT}(\mathbb{S},t_\text{stop})=x^*$}
\label{alg:elect}
\begin{algorithmic}[1]
\Require Source image $I$, Edit instruction $C_T$, Candidate seed set $\mathbb{S}$, stopping timestep $t_{\text{stop}}$, instruction-guided denoiser $\epsilon_\theta$, VAE encoder $\mathcal{E}$ and decoder $\mathcal{D}$
\Ensure Best edited image $x^*$
\State $z\gets \mathcal{E}(I)$
\State Sample $z_T^i\sim \mathcal{N}(0,I)$ with seed $i$ for all $i\in\mathbb{S}$
\For{$t = T \to t_{\text{stop}}+1$} \Comment{Denoise until stopping time}
    \For{$i \in \mathbb{S}$}
        \State $z_{t-1}^i \gets \texttt{Denoise}(z_t^i,t,I,C_{T})$ 
    \EndFor
\EndFor
\For{$i \in \mathbb{S}$}
    \State $S^\text{BIS}(i,t_\text{stop})\gets S^\text{BIS}(i,t_\text{stop}\mid \mathbb{S},\epsilon_\theta,I,C_T)$ as in \eqref{eq:S^BIS}
\EndFor
\State $i^* \gets \arg\min_{i \in \mathbb{S}} S^\text{BIS}(i, t_{\text{stop}})$ \Comment{Select best seed}
\For{$t = t_{\text{stop}} \to 1$} \Comment{Continue denoising $i^*$}
    \State $z^{i^*}_{t-1} \gets \texttt{Denoise}(z^{i^*}_{t},t,I,C_T)$
\EndFor
\State \Return $x^* \gets \mathcal{D}(z^{i^*}_{0})$ \Comment{Final edited image}
\end{algorithmic}
\end{algorithm}

ELECT is designed to efficiently select the best candidate for image editing while reducing computational costs. As shown in \cref{fig:overall_pipeline}, ELECT evaluates multiple candidates early in the denoising process, eliminating suboptimal ones at an early timestep $t_\text{stop}$ before completing inference. ELECT ensures that only the most promising sample is fully denoised, balancing efficiency and accuracy in generative image editing. 
Following DDIM \cite{DDIM_song2020denoising}, $\texttt{Denoise}(z_t,t,I,C_T)=\sqrt{\alpha_{t-1}}\hat{z}_0+\sqrt{1-\alpha_{t-1}}\cdot \epsilon_\theta(z_t,t,I,C_T)$ in \cref{alg:elect}. For rectified flow, $\texttt{Denoise}(z_t,t,I,C_T)=z_t-v_\theta (z_t,t,I,C_T)\cdot \Delta t$ where the time domain is $t\in[0,1]$.

\begin{figure*}[t]
    \centering
    \includegraphics[width=0.7\textwidth]{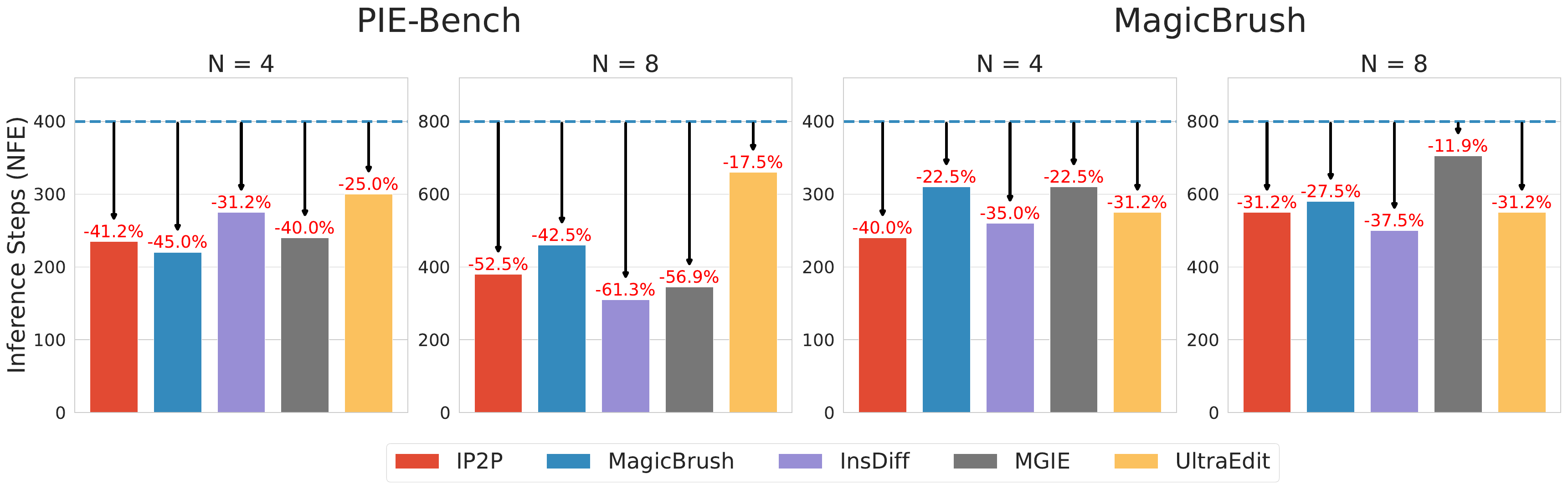}
    \vspace{-0.3cm}
    \caption{\textbf{Efficiency Comparison of ELECT vs. Best of N by $S^{\text{BIS}}$ for Comparable Performance}. Comparing the time cost (NFE) and reduction rate of ELECT with Best of N by $S^\text{BIS}$ (\blue{blue line}), which undergoes all denoising steps, for comparable performance on various models \cite{IP2P_brooks2023instructpix2pix, MagicBrush_NEURIPS2023_64008fa3, InstructDiff_Geng23instructdiff,MGIE_fu2024guiding,UltraEdit_zhao2024ultraedit} (Left: PIE-Bench \cite{PnP_Inversion_ju2023direct}, Right: MagicBrush \cite{MagicBrush_NEURIPS2023_64008fa3} test set). The comparison is based on evaluations with similar Background MSE values. Since MSE values are continuous, we focus on conditions where ELECT's performance marginally excels within an error range of 1e-5. NFE is determined by ELECT's number of seeds and stopping timesteps. In the MagicBrush test set, although the performance gain is not as significant as in PIE-bench due to factors such as image complexity and instructional difficulty, a notable improvement in efficiency is still demonstrated.}
    \label{fig:efficiency}
    \vspace{-0.5cm}
\end{figure*}

\subsection{ELECT for Instruction Prompt Selection}

We found that seed selection improves image editing performance, but its effectiveness plateaus as the number of seeds increases. For out-of-distribution instruction prompts, seed selection alone won't yield a successful edit. In such cases, modifying the instruction prompt results in improvements (see \cref{fig:seed_variance}).

To incorporate a Multimodal Large Language Model (MLLM) \cite{gpt4o,GPT4V_yang2023dawn} into the ELECT pipeline, we introduce a new evaluation metric inspired by prior work \cite{ku2024imagenhub, Ku2023VIEScoreTE}. This metric classifies edits based on two sub-metrics: Instruction Following and Background Consistency. These sub-metrics independently assess alignment with the instruction and preservation of unedited areas, giving scores of $0,0.5,$ or $1$. If either score is 0, the edit fails, prompting an MLLM to generate alternative instruction prompts while preserving meaning. The ELECT pipeline is then applied to select the best prompt early on \cref{app:prompt_selection}.

\section{Experiments}
\label{sec:experiments}
\begin{figure*}[tp]
    \centering
    \includegraphics[width=0.9\textwidth]{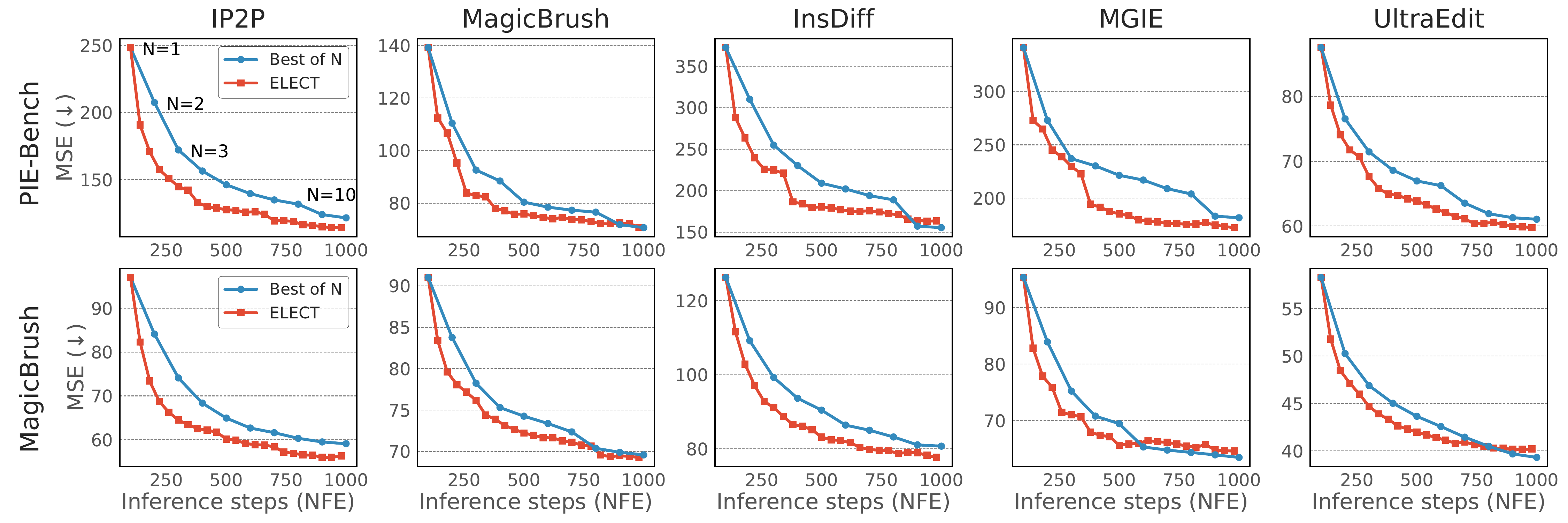}
    \vspace{-0.2cm}
    \caption{\textbf{Quantitative Comparison of MSE between ELECT and Best of N by $S^\text{BIS}$.} Performance trend (MSE$_{\times10^4}$) with respect to the number of function evaluations (NFE), evaluated on two datasets with $t_\text{stop}=60$ (Top: PIE-Bench \cite{PnP_Inversion_ju2023direct}, Bottom: MagicBrush \cite{MagicBrush_NEURIPS2023_64008fa3} test set). Results show consistent performance improvements and enhanced efficiency across various models and datasets. By evaluating more seeds within the same NFE budget, ELECT achieves superior overall performance, surpassing the Pareto front of Best of N by $S^\text{BIS}$.}
    \label{fig:main}
    \vspace{-0.2cm}
\end{figure*}

\subsection{Setup}
We validate our method for instruction-guided image editing by measuring its effectiveness in both instruction following and background consistency.

\textbf{Baselines.} We compare ELECT against five diffusion-based editing models\textemdash InstructPix2Pix (IP2P) \cite{IP2P_brooks2023instructpix2pix}, MagicBrush \cite{MagicBrush_NEURIPS2023_64008fa3}, InstructDiffusion (InsDiff) \cite{InstructDiff_Geng23instructdiff}, MGIE \cite{MGIE_fu2024guiding}, and UltraEdit \cite{UltraEdit_zhao2024ultraedit}\textemdash operating under a constrained setting without ground-truth masks or prompts. Additionally, we introduce \textbf{Best of N by $S^\text{BIS}$},  which selects the best output via the Background Inconsistency Score \eqref{eq:S^BIS} after full inference (i.e., 100 denoising steps), serving as a direct comparison when ELECT's stopping step is set to zero.

\textbf{Benchmarks and Metrics.} Experiments are conducted on PIE-Bench \cite{PnP_Inversion_ju2023direct}, covering 9 editing scenarios with 700 real and synthetic images, and the MagicBrush \cite{MagicBrush_NEURIPS2023_64008fa3} test set, a manually-annotated real image dataset containing around 560 images. Each dataset provides a source image, edit instruction, and GT foreground mask which is used only for evaluation. Performance is evaluated via CLIPScore \cite{hessel2021clipscore} for \textit{instruction following} and PSNR, MSE, SSIM \cite{Wang2004ImageQA}, and LPIPS \cite{Zhang2018TheUE} for \textit{background consistency}. We also report VIEScore \cite{Ku2023VIEScoreTE}, a human-aligned metric assessing overall edit quality. For the full implementation detail, refer to \cref{app:detailed_experimental}.

\subsection{Effect of Seed Selection with ELECT}
\begin{figure}[tp]
    \centering
    \includegraphics[width=\linewidth]{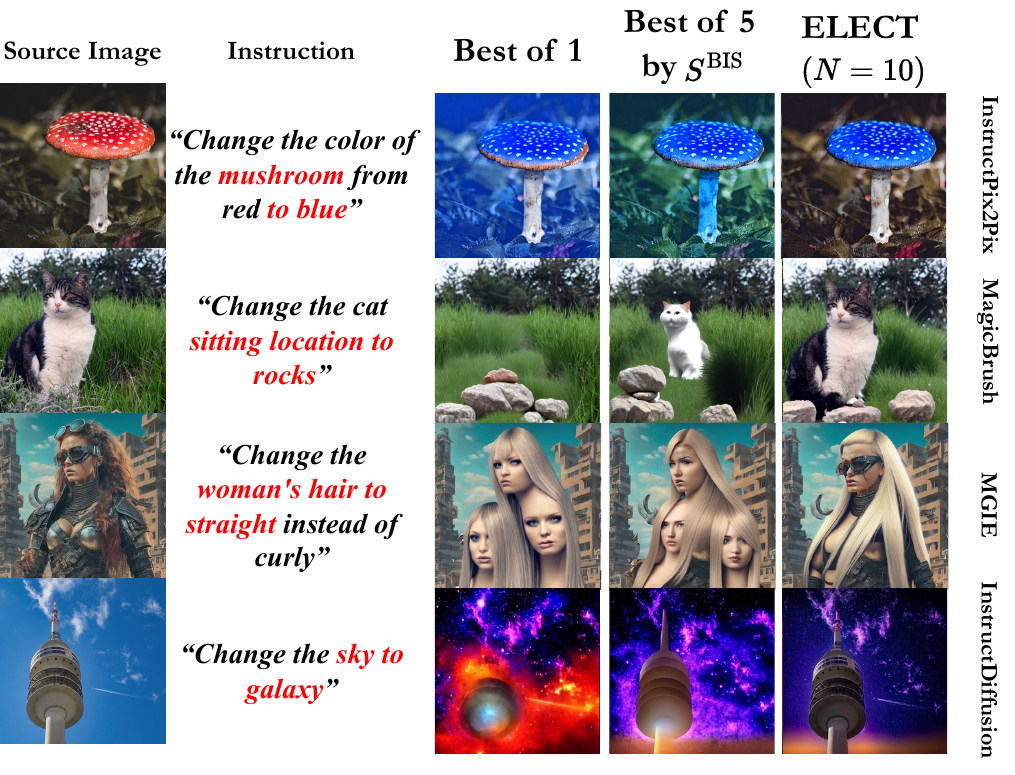}
    \vspace{-0.6cm}
    \caption{\textbf{Qualitative Result for Seed Selection}: When a single-seed image suffers from severe distortion, examining multiple outputs enables the selection of a less distorted sample. Since ELECT explores more seeds than Best of 5 by $S^\text{BIS}$, this effect is further amplified, leading to better overall image selection.
}
    \label{fig:qualitative}
    \vspace{-0.3cm}
\end{figure}

We evaluate ELECT in terms of both performance and efficiency (denoising steps required) for instruction-guided image editing. As shown in \cref{tab:comparison}, multi-seed strategies significantly outperform single-seed evaluation (Vanilla). ELECT achieves the best results, consistently surpassing all baselines across all metrics, particularly when matched to the time complexity of Best of 5 by $S^\text{BIS}$. \cref{fig:qualitative} provides a qualitative comparison across multiple models, comparing Best of 1 (vanilla), Best of 5 with $S^\text{BIS}$, and ELECT. Single-seed outputs often exhibit excessive distortion, while selecting a seed with higher background consistency (as in Best of 5) reduces this issue. However, evaluating more seeds within the same computational budget allows ELECT to further minimize unnecessary background modifications. Additional qualitative results are available in \cref{app:additional_qualitative_results}.

\textbf{Performance.} \cref{fig:main} illustrates consistent performance improvements and enhanced efficiency across various models and datasets. Background Consistency is measured using Mean Squared Error (MSE) (y-axis), while the Number of Function Evaluations (NFEs) (x-axis) represents the inference steps in diffusion models, serving as a proxy for computational cost. By evaluating more seeds within the same NFE budget, ELECT achieves superior overall performance, surpassing the Pareto front of Best of N by $S^\text{BIS}$ in the graph. As the number of seeds (N) increases, performance improves but eventually plateaus, typically converging within 1000 NFEs. Notably, for MGIE and MagicBrush, this saturation point is often reached more quickly, depending on the dataset.

\textbf{Efficiency.} Our method demonstrates superior efficiency across all models (\cref{fig:efficiency}). ELECT required less than 50\% of the NFEs used by Best of N (N = 8) in PIE-Bench, except for UltraEdit, reducing time costs by 36.2\% on average. In the MagicBrush test set, where images and instructions are more complex, performance gains from increasing the number of seeds were smaller but still meaningful, highlighting the efficiency of ELECT and the effectiveness of Best of N.

\begin{table*}[ht]
    \caption{\textbf{Comparison of Different Selection Methods on PIEBench.} We conducted a quantitative evaluation from the perspectives of Instruction Following (IF) and Background Consistency (BC) and utilized the Semantic Consistency component of VIEScore, a metric based on MLLM that exhibits strong human alignment across these two perspectives. \textbf{ELECT} selects one seed at $t_{\text{stop}}=60$ from $N=11$ and determines the best seed by computing the Background Inconsistency Score (BIS). This result was compared with the baseline using a single seed \textbf{(Best of 1: Vanilla)} and a fair comparison where the best result was selected from outputs based on BIS \textbf{(Best of N)}.}
\vspace{-0.2cm}
\resizebox{\textwidth}{!}{
    \begin{tabular}{l|c|cccc|c|ccc|c}
        \toprule
        \multirow{2}{*}{Model} & \multirow{2}{*}{\makecell{ Seed Selection \\ Method}} & \multicolumn{4}{c|}{BC} & \multicolumn{1}{c|}{IF} & \multicolumn{3}{c|}{VIEScore (Semantic Consistency) ($\uparrow$)} & \multirow{2}{*}{\makecell{Time-complexity \\ (NFE)}} \\
        \cmidrule{3-10}
        & & $\text{MSE}_{\times10^4}$ ($\downarrow$) & $\text{LPIPS}_{\times10^3}$ ($\downarrow$) & PSNR ($\uparrow$) & $\text{SSIM}_{\times10^2}$ ($\uparrow$) & CLIP-T ($\uparrow$) & BC & IF & min(BC, IF)  \\
        \midrule
    \multirow{3}{*}{IP2P \cite{IP2P_brooks2023instructpix2pix}} & Vanillla  & 248.493 & 162.414 & 20.734 & 75.976 & 24.380 & 6.017 & 4.151 & 3.430 & 100 \\
     & best of N by $S^\text{BIS}$ & \underline{146.151} & \underline{113.827} & \underline{22.953} & \underline{80.132} & \underline{24.682} & \underline{6.621} & \underline{4.210} & \underline{3.570} & 500  \\
     & \cellcolor[HTML]{ECF4FF}\textbf{ELECT} & \cellcolor[HTML]{ECF4FF}\textbf{127.481} & \cellcolor[HTML]{ECF4FF}\textbf{103.338} & \cellcolor[HTML]{ECF4FF}\textbf{23.329} & \cellcolor[HTML]{ECF4FF}\textbf{80.902} & \cellcolor[HTML]{ECF4FF}\textbf{24.974} & \cellcolor[HTML]{ECF4FF}\textbf{6.824} & \cellcolor[HTML]{ECF4FF}\textbf{4.252} & \cellcolor[HTML]{ECF4FF}\textbf{3.667} & \cellcolor[HTML]{ECF4FF}500  \\
    \midrule
    \multirow{3}{*}{MagicBrush \cite{MagicBrush_NEURIPS2023_64008fa3}} & Vanillla & 139.178 & 77.222 & 24.833 & 82.839 & 24.628 & 5.887 & 4.699 & 3.986 & 100 \\
     & best of N by $S^\text{BIS}$ & \underline{80.406} & \underline{59.869} & \textbf{26.253} & \underline{84.615} & \underline{25.000} & \underline{6.191} & \underline{4.760} & \underline{4.133} & 500  \\
     & \cellcolor[HTML]{ECF4FF}\textbf{ELECT} & \cellcolor[HTML]{ECF4FF}\textbf{75.901} & \cellcolor[HTML]{ECF4FF}\textbf{59.104} & \cellcolor[HTML]{ECF4FF}\underline{26.133} & \cellcolor[HTML]{ECF4FF}\textbf{84.706} & \cellcolor[HTML]{ECF4FF}\textbf{25.067} & \cellcolor[HTML]{ECF4FF}\textbf{6.261} & \cellcolor[HTML]{ECF4FF}\textbf{4.881} & \cellcolor[HTML]{ECF4FF}\textbf{4.224} & \cellcolor[HTML]{ECF4FF}500  \\
        \midrule
    \multirow{3}{*}{InsDiff \cite{InstructDiff_Geng23instructdiff}} & Vanillla & 372.465 & 154.041 & 20.251 & 75.530 & 24.091 & 5.420 & 4.179 & 3.534 & 100 \\
     & best of N by $S^\text{BIS}$ & \underline{208.998} & \underline{108.448} & \underline{22.753} & \underline{79.962} & \underline{24.543} & \underline{5.750} & \underline{4.424} & \underline{3.767} & 500  \\
     & \cellcolor[HTML]{ECF4FF}\textbf{ELECT} & \cellcolor[HTML]{ECF4FF}\textbf{180.524} & \cellcolor[HTML]{ECF4FF}\textbf{104.518} & \cellcolor[HTML]{ECF4FF}\textbf{22.849} & \cellcolor[HTML]{ECF4FF}\textbf{80.026} & \cellcolor[HTML]{ECF4FF}\textbf{24.746} & \cellcolor[HTML]{ECF4FF}\textbf{5.871} & \cellcolor[HTML]{ECF4FF}\textbf{4.545} & \cellcolor[HTML]{ECF4FF}\textbf{3.817} & \cellcolor[HTML]{ECF4FF}500  \\
        \midrule
    \multirow{3}{*}{MGIE \cite{MGIE_fu2024guiding}} & Vanillla & 341.418 & 145.512 & 21.164 & 77.312 & 24.438 & 5.640 & 4.409 & 3.679 & 100 \\
     & best of N by $S^\text{BIS}$ & \underline{221.394} & \underline{111.690} & \underline{23.183} & \underline{80.626} & \underline{24.603} & \underline{6.226} & \underline{4.560} & \underline{3.903} & 500  \\
     & \cellcolor[HTML]{ECF4FF}\textbf{ELECT} & \cellcolor[HTML]{ECF4FF}\textbf{185.077} & \cellcolor[HTML]{ECF4FF}\textbf{102.536} & \cellcolor[HTML]{ECF4FF}\textbf{23.605} & \cellcolor[HTML]{ECF4FF}\textbf{81.337} & \cellcolor[HTML]{ECF4FF}\textbf{24.727} & \cellcolor[HTML]{ECF4FF}\textbf{6.265} & \cellcolor[HTML]{ECF4FF}\textbf{4.592} & \cellcolor[HTML]{ECF4FF}\textbf{3.953} & \cellcolor[HTML]{ECF4FF}500  \\
        \midrule
    \multirow{3}{*}{UltraEdit \cite{UltraEdit_zhao2024ultraedit}} & Vanillla & 87.544 & 115.365 & 22.929 & 79.859 & 25.197 & 5.889 & 5.500 & 4.466 & 100 \\
     & best of N by $S^\text{BIS}$ & \underline{66.958} & \underline{96.269} & \underline{24.374} & \underline{83.004} & \textbf{25.379} & \underline{6.279} & \underline{5.571} & \underline{4.681} & 500  \\
     & \cellcolor[HTML]{ECF4FF}\textbf{ELECT} & \cellcolor[HTML]{ECF4FF}\textbf{63.847} & \cellcolor[HTML]{ECF4FF}\textbf{92.312} & \cellcolor[HTML]{ECF4FF}\textbf{24.492} & \cellcolor[HTML]{ECF4FF}\textbf{83.649} & \cellcolor[HTML]{ECF4FF}25.362 & \cellcolor[HTML]{ECF4FF}\textbf{6.369} & \cellcolor[HTML]{ECF4FF}\textbf{5.590} & \cellcolor[HTML]{ECF4FF}\textbf{4.695} & \cellcolor[HTML]{ECF4FF}500 \\
        \bottomrule
    \end{tabular}
    }
    \vspace{-0.5cm}
    \label{tab:comparison}
\end{table*}

\subsection{Effect of Instruction Prompt Selection with ELECT}

\begin{figure}[tp]
    \centering
    \includegraphics[width=\linewidth]{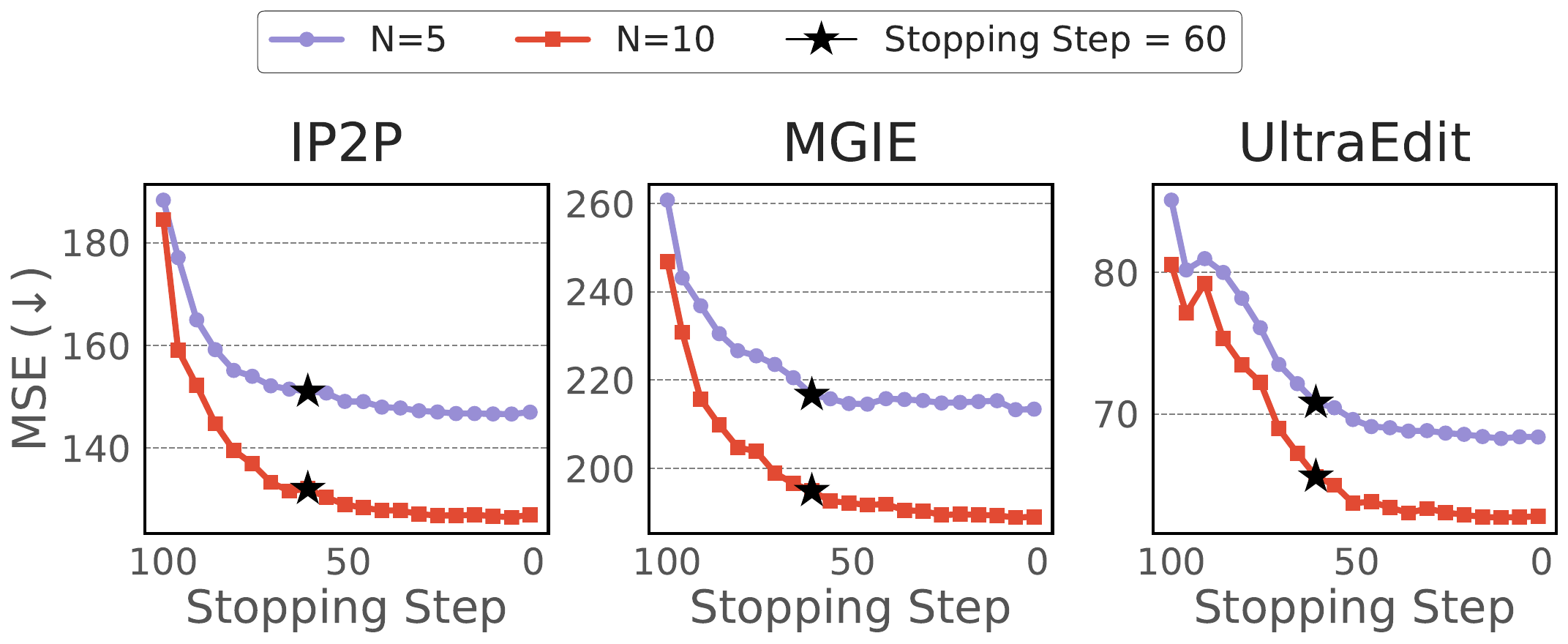}
    \vspace{-0.6cm}
    \caption{\textbf{ELECT performance variation with respect to stopping timestep ($t_\text{stop}$) with fixed number of seeds}. This graph shows that performance improves as the stopping denoising step increases, eventually converging around around $t_\text{stop} = 70$ for most models. In contrast, UltraEdit, as a Rectified Model, exhibits minimal change in noise ratio at very early steps, making it meaningful to select a stopping point after approximately $t_\text{stop}=60$. Full analysis is in the supplementary materials.}
    \label{fig:stopping_steps}
    \vspace{-0.3cm}
\end{figure}

\cref{fig:prompt_selection_example} illustrates that seed selection alone is insufficient to consistently generate high-quality samples. In many cases, the model often fails to properly reflect the input conditions, consistently producing either severely distorted images or no meaningful edits at all. To address this, introducing prompt variants provides diverse signals, increasing the likelihood that the model successfully applies the desired edits to the given image.

\begin{figure}[ht]
    \centering
    \vspace{-0.3cm}
    \includegraphics[width=\columnwidth]{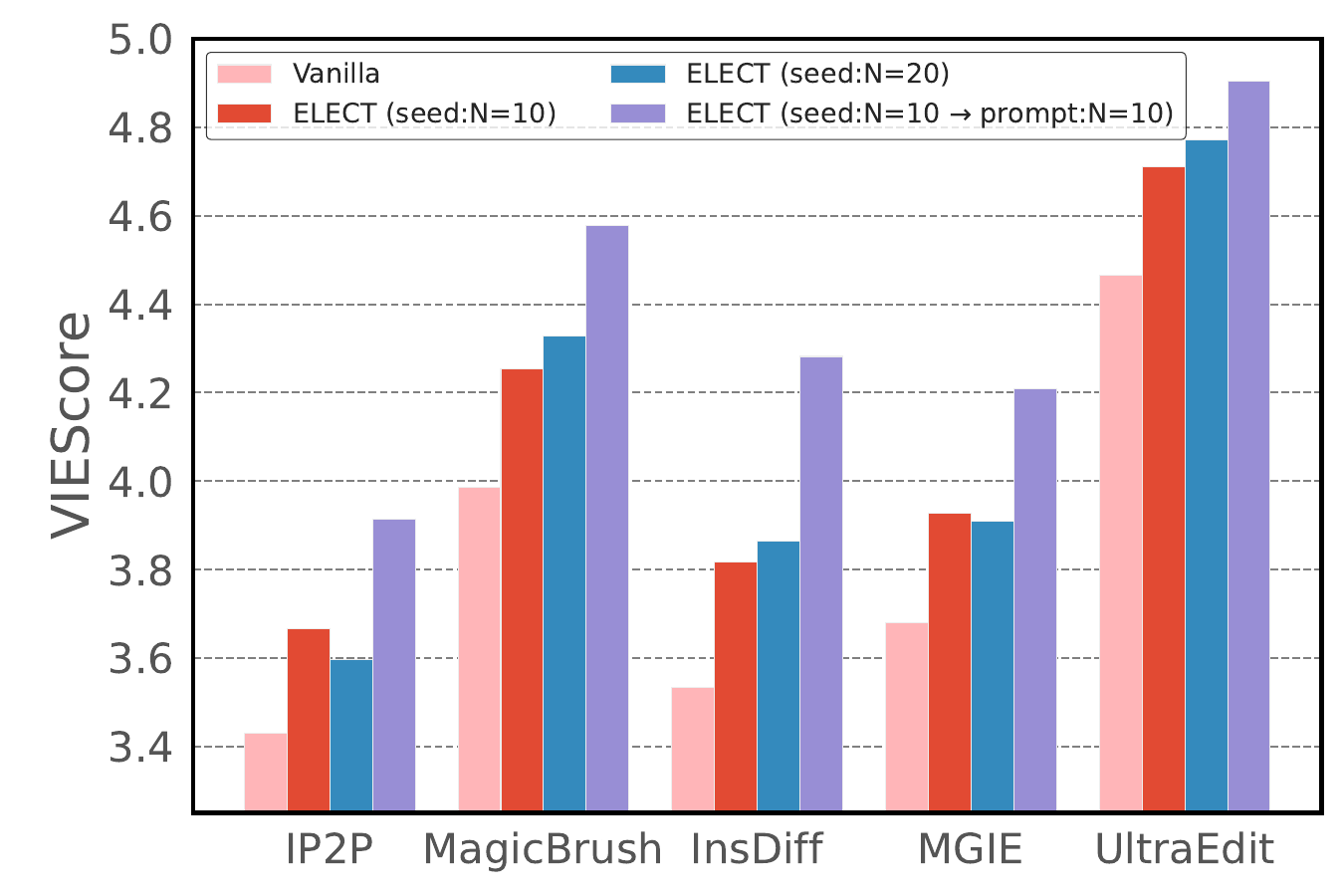}
    \vspace{-0.6cm}
    \caption{Comparison of VIEScore (Semantic Consistency) across three settings\textemdash Single Seed sampling, Seed Selection, and Prompt Selection after Seed Selection\textemdash shows that adding prompt variance ($N=10$ after Seed Selection) improves editing outcomes more than simply increasing seed candidates ($N=20$).}
    \vspace{-0.2cm}
    \label{fig:prompt_selection_viescore}
\end{figure}

These failure cases are further analyzed in \cref{fig:prompt_selection_viescore}, which presents a comparison of VIEScore metrics before and after prompt selection for samples initially classified as failures following seed selection under ELECT. Simply increasing the number of seed candidates leads to performance saturation or over-optimization, where no further improvements are observed in evaluation metrics. However, across all models, prompt selection effectively overcomes this saturation, resulting in a significant increase in VIEScore. A comprehensive quantitative comparison of all evaluation metrics and more qualitative results are provided in \cref{app:additional_qualitative_results}.

\subsection{Ablation Study}

\textbf{Choice of $t_\text{stop}$}. In ELECT, the hyperparameter $t_{\text{stop}}$ controls when denoising stops and candidates are compared, balancing efficiency and performance. More denoising steps improve alignment with the final output, approaching Best of N performance but increasing time complexity. Conversely, stopping too early results in noisy candidates, making stable scoring difficult. Theoretically, the signal-to-noise ratio (SNR) reaches 1 after 20 steps, allowing reliable comparisons beyond this point. As shown in (\cref{fig:stopping_steps}), empirically, most diffusion models achieve stable performance after $t_\text{stop}=70$, and UltraEdit (Rectified Flow) requires at least $t_\text{stop}=60$. 

To address model/sample variability in $t_{\text{stop}}$, we also explored a method that determines the early step for comparison adaptively rather than using a fixed step. We found that this performance convergence trend closely resembles the point at which the Score converges with respect to the timestep. Based on this observation, we attempted to automatically identify this point for each sample (see Supplementary) (\cref{fig:result_grid}). In practice, this approach consistently and stably improved performance across all models while also enhancing efficiency.

\begin{figure}[tp]
    \centering
    \includegraphics[width=\linewidth]{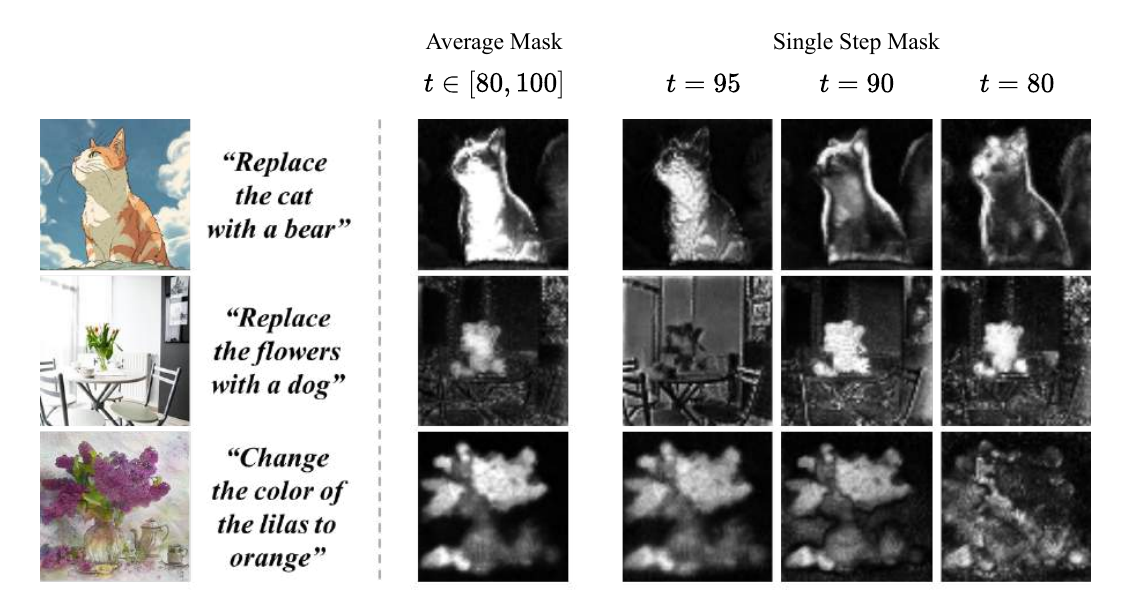}
    \vspace{-0.7cm}
    \caption{\textbf{Extracted masks at different timesteps.} The right three columns show masks extracted at individual denoising steps ($t=95, 90$, and $80$) for IP2P \cite{IP2P_brooks2023instructpix2pix}. The leftmost column of masks shows the averaged mask over $t\in[80, 100]$, which consistently yields more reliable results across diverse cases}
    \vspace{-0.5cm}
    \label{fig:mask_ablation}
\end{figure}

\textbf{Mask Extraction}.As shown in \cref{fig:mask_ablation}, the timestep at which the primary editing region is captured in the relevance map varies across samples, even within the same model. Some images require 10–20 denoising steps for a well-defined map, while others capture fine details early on. After 20 steps, masks focus on high-frequency regions, such as edges. To address this, we propose using an average mask over timesteps $t \in [0, 20]$, to provide a stable method for detecting editing regions across samples. This approach eliminates the need for a fixed timestep, as optimal denoising steps differ per sample. We replace $M_{t}^{\text{mean}}$ in Equation (\ref{eq:S^BIS}) with the expectation $\mathbb{E}_{t \sim \mathcal{U}[80,100]}[M_{t}^{\text{mean}}]$, improving mask extraction robustness.

\section{Conclusion}
\label{sec:conclusion}

In this work, we introduced ELECT, a zero-shot framework that enhances instruction-guided image editing by selecting seeds that preserve background consistency while modifying the foreground. ELECT establishes a new multiple-seed editing baseline, achieving Best-of-N performance without supervision.
Experiments show that ELECT reduce computational costs by 41\% on average (up to 61\%) while improving background consistency and instruction adherence. Additionally it integrates with editing pipelines and MLLMs for joint seed and prompt selection, further enhancing results. By eliminating reliance on external verifiers and reducing computation, ELECT provides an efficient and practical solution for diffusion-based image editing. 

\paragraph{Limitations}
Our relative score technique selects candidates based on self-supervised background consistency. While this may sometimes over-optimize for image preservation, such cases are rare and do not significantly impact performance. ELECT improves instruction adherence, as shown by CLIPScore and VIEscore, demonstrating that over-optimization is not a major concern.

\clearpage

\section*{Acknowledgments}
This work was supported by multiple grants from the Institute of Information \& Communications Technology Planning \& Evaluation (IITP), funded by the Korean government (MSIT): RS-2024-00457882 (AI Research Hub Project), RS-2025-25442338 (AI Star Fellowship Support Program, Seoul National University), RS-2025-02305581, and RS2021-II211343 (Artificial Intelligence Graduate School Program, Seoul National University). It was also supported by a grant from the Korea Health Technology R\&D Project through the Korea Health Industry Development Institute (KHIDI), funded by the Ministry of Health \& Welfare, Republic of Korea (RS-2025-02307233), and by the National Research Foundation of Korea (NRF) grants funded by MSIT: RS-2023-00209060 (A Study on Optimization and Network Interpretation Method for Large-Scale Machine Learning), RS-2023-00242443, and RS-2023-00282907. This research was also conducted as part of the "AI Media and Cultural Enjoyment Expansion" Project, supported by the Ministry of Science and ICT and the National IT Industry Promotion Agency (NIPA) in 2025.

{\small
\bibliographystyle{ieeenat_fullname}
\bibliography{_main}
}

\clearpage \appendix 

\section{Comparison with Existing Work}

We provide a table comparing our work with previous image editing studies in \cref{tab:baseline_comparison}

\section{Detailed Experimental Setup}
\label{app:detailed_experimental}

Our experiment evaluates the effectiveness and efficiency of our candidate selection method for image editing, focusing on its ability to follow user instructions while maintaining the source image's visual fidelity.


\textbf{Baselines}. We establish 5 diffusion-based instruction-guided image editing models as baselines. All models operate under a constrained setting where they take only the source image and user instruction as inputs, without access to ground-truth masks or source/target prompts. The instruction-guided image editing models considered in this work include InstructPix2Pix (IP2P) \cite{IP2P_brooks2023instructpix2pix}, MagicBrush \cite{MagicBrush_NEURIPS2023_64008fa3}, InstructDiffusion (InsDiff) \cite{InstructDiff_Geng23instructdiff}, MGIE \cite{MGIE_fu2024guiding}, and UltraEdit \cite{UltraEdit_zhao2024ultraedit}. Among them, UltraEdit is a fine-tuned model based on Stable Diffusion 3, demonstrating that our method can also enhance the performance of Rectified-Flow models effectively.

Since there is no existing method for seed selection in image editing, we compare our approach, \textbf{ELECT}, with new baseline '\textbf{Best of N by $S^{\text{BIS}}$}' (hereafter referred to as \textbf{Best of N}), which selects the best output via Background Inconsistency Score (BIS) after evaluating all generated samples. This is equivalent to the ELECT algorithm when $t_\text{stop}=0$. While Best of N compares outputs after running the full 100 denoising steps for each initial noise, our method selects the best seed after evaluating only 40 denoising steps. 

\textbf{Benchmarks}. We use two well-known benchmarks to evaluate the image editing task. First, PIE-Bench \cite{PnP_Inversion_ju2023direct} provides a test set covering 9 different editing scenarios and includes data from both real and AI-generated image domains, consisting of 700 images. Second, the MagicBrush test set \cite{MagicBrush_NEURIPS2023_64008fa3}, consists of a manually-annotated dataset that allows evaluation on real images and scenarios, containing around 560 images. Each dataset provides a source image, editing instruction, and foreground object mask, where the mask is used only for metric evaluation.

\textbf{Metrics}. We evaluate image editing performance using two key objectives: (1) Instruction Following and (2) Background Consistency. Instruction Following is measured with CLIPScore \cite{hessel2021clipscore}, assessing semantic similarity between the edited image and target caption in CLIP's \cite{radford2021learning} embedding space. For background consistency, we evaluate the visual fidelity of the edited image relative to the source image using PSNR, MSE, SSIM \cite{Wang2004ImageQA}, and LPIPS \cite{Zhang2018TheUE}, leveraging the dataset’s ground-truth mask. We also use VIEScore \cite{Ku2023VIEScoreTE} (0-10), which aligns with human preferences and combine both objectives via MLLM-based evaluation. To gain a more detailed perspective, we separately record the Instruction Following score and Background Consistency score, which constitute the Semantic Consistency (SC) score within VIEScore.

\begin{figure}[H]
    \centering
    \begin{subfigure}{0.9\columnwidth}
        \centering
        \includegraphics[width=\linewidth]{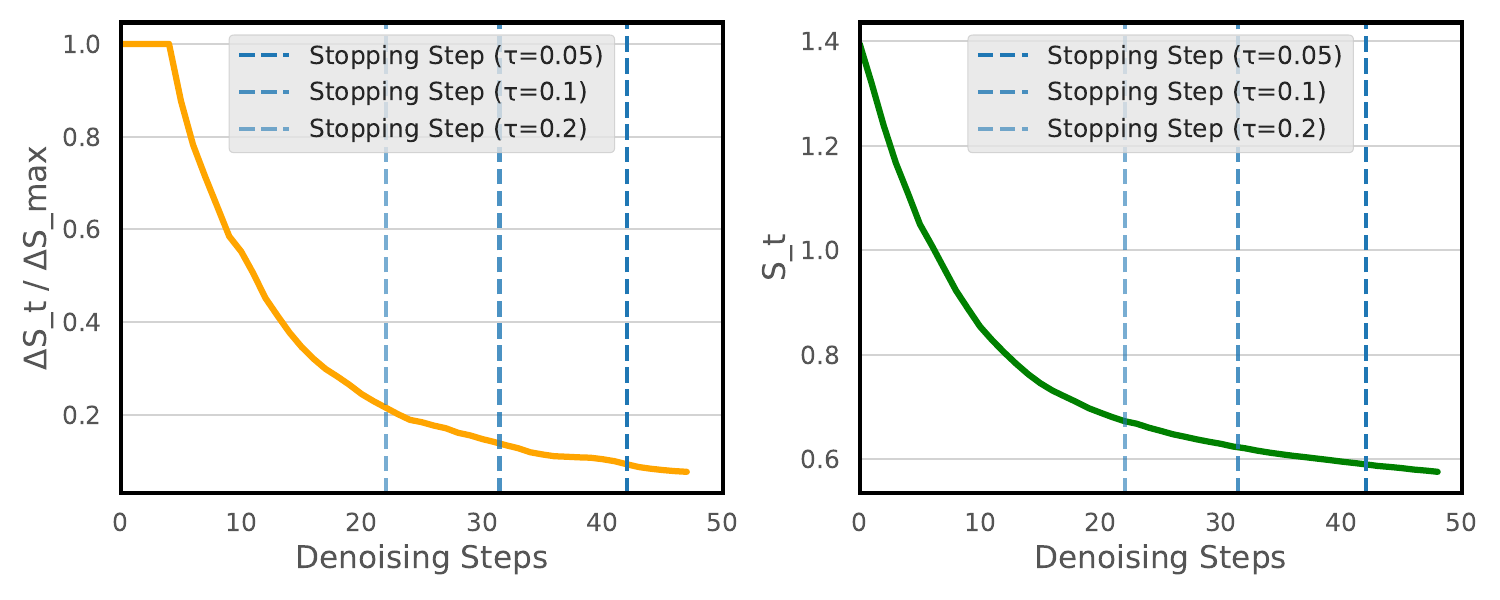}
        \caption{InstructPix2Pix}
        \label{fig:ip2p_score}
    \end{subfigure}

    \vspace{0em}
    \begin{subfigure}{0.9\columnwidth}
        \centering
        \includegraphics[width=\linewidth]{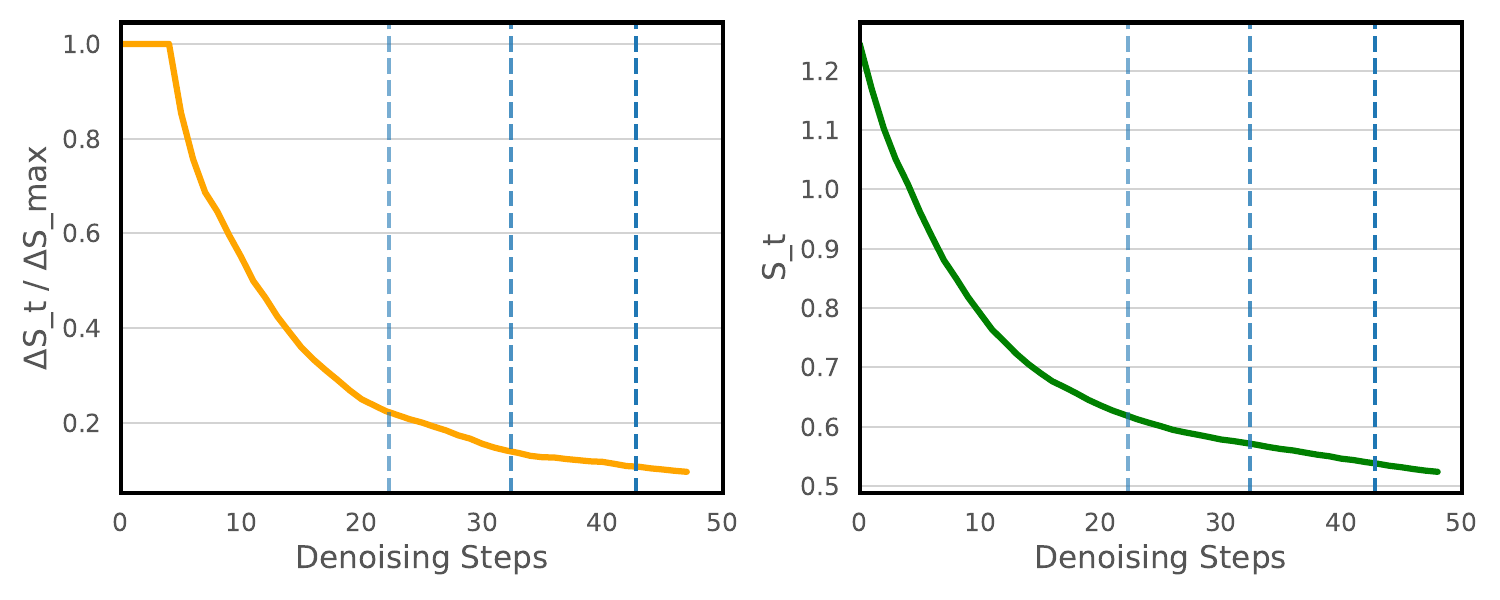}
        \caption{MagicBrush}
        \label{fig:mb_score}
    \end{subfigure}
    
    \vspace{0em}
    \begin{subfigure}{0.9\columnwidth}
        \centering
        \includegraphics[width=\linewidth]{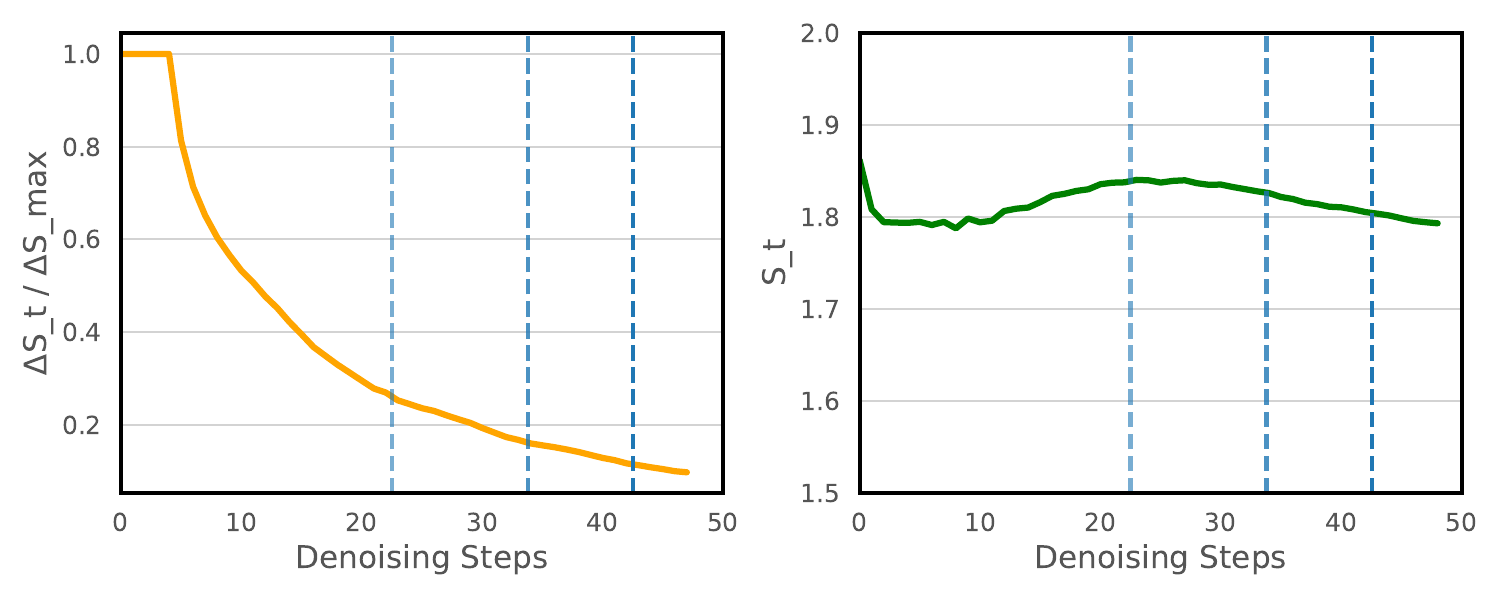}
        \caption{InstructDiffusion}
        \label{fig:id_score}
    \end{subfigure}

    \vspace{0em}
    \begin{subfigure}{0.9\columnwidth}
        \centering
        \includegraphics[width=\linewidth]{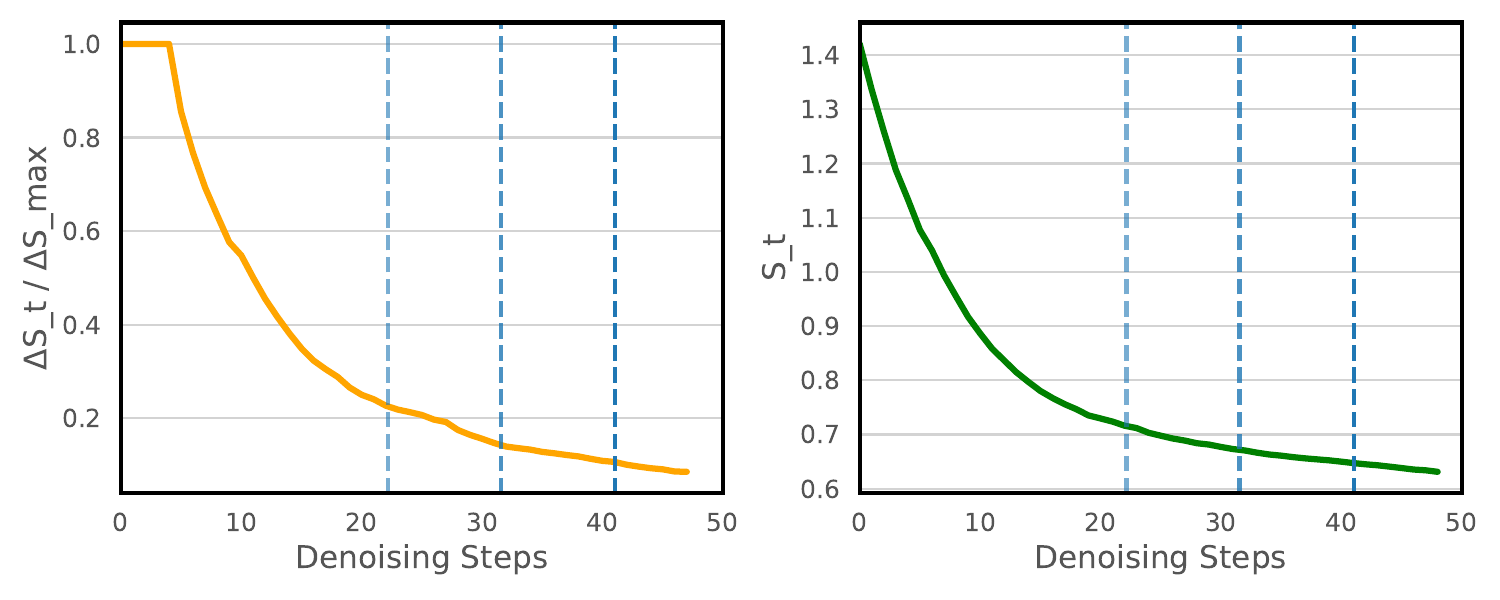}
        \caption{MGIE}
        \label{fig:mgie}
    \end{subfigure}
    
    \vspace{0em}

    \begin{subfigure}{0.9\columnwidth}
        \centering
        \includegraphics[width=\linewidth]{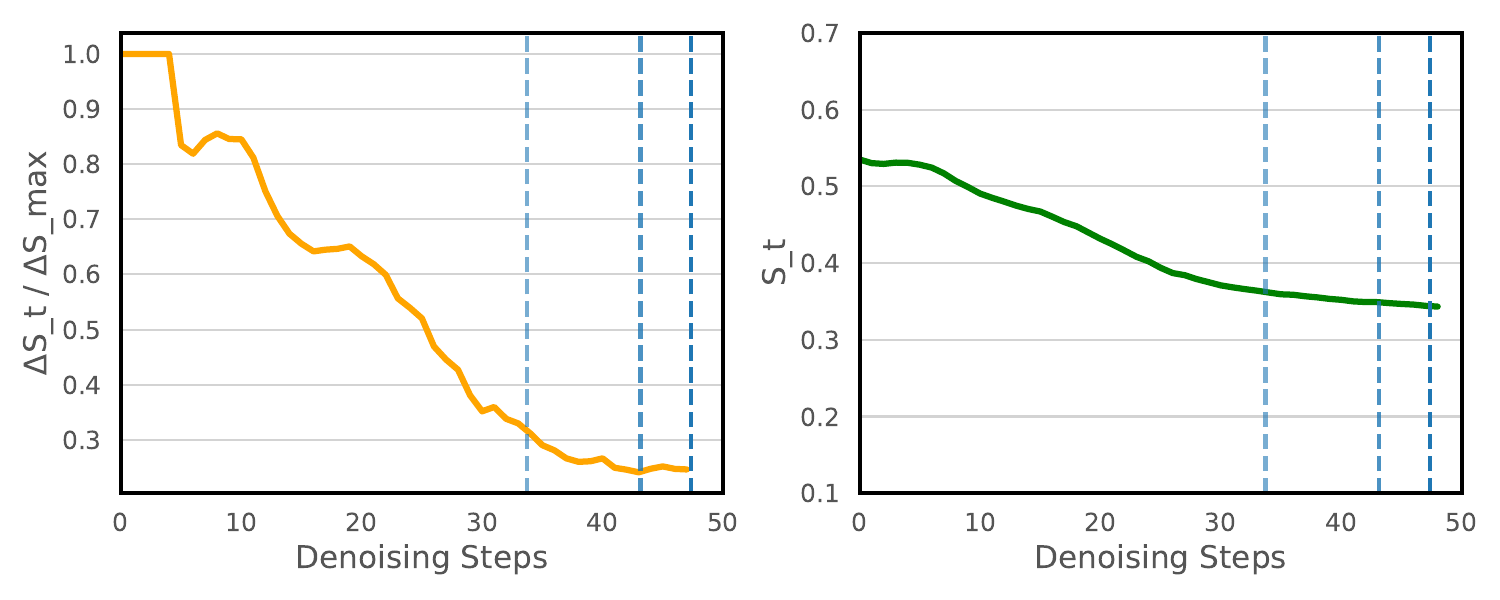}
        \caption{UltraEdit}
        \label{fig:ue_score}
    \end{subfigure}

    \caption{Experimental motivation and implementation of the diminishing delta criterion, which halts the denoising process once the delta score falls below a threshold defined as $\tau \cdot \Delta S_\text{max}$. The graphs illustrate the evolution of the score and delta score over timesteps, with convergence behavior observed for small $\tau$. A moving average over 5 timesteps is applied to enhance robustness against noise. }
    \label{fig:result_grid}
\end{figure}

\begin{table*}
\caption{\textbf{Comparison of Methods Addressing Background Inconsistency in Text-guided Image Editing}. Our method is the first to introduce optimal seed selection for instruction-guided editing and uniquely enables MLLM-based instruction prompt selection, which is absent in existing approaches. Unlike prior methods, our ELECT framework achieves these capabilities without requiring external segmentation models or source/target prompt pairs.}
\vspace{-0.3cm}
\resizebox{\textwidth}{!}{
\begin{tabular}{lllllllll}
\hline
\rowcolor[HTML]{FFFFFF} & Ours & WYS \cite{Watch_Your_Steps_mirzaei2025watch} & ZONE \cite{ZONE_li2024zone} & MagicBrush \cite{MagicBrush_NEURIPS2023_64008fa3} & UltraEdit \cite{UltraEdit_zhao2024ultraedit} & DirectInversion \cite{PnP_Inversion_ju2023direct} & InfEdit \cite{InfEdit_xu2023infedit} & NTI \cite{NTI_2023_CVPR}, PTI \cite{PTI_dong2023prompt} \\
\hline
Optimal Seed Selection & \textcolor{green}{\cmark} & \textcolor{red}{\xmark} & \textcolor{red}{\xmark} & \textcolor{red}{\xmark} & \textcolor{red}{\xmark} & \textcolor{red}{\xmark} & \textcolor{red}{\xmark} & \textcolor{red}{\xmark} \\
Optimal Prompt Selection/Tuning & \textcolor{green}{\cmark} & \textcolor{red}{\xmark} & \textcolor{red}{\xmark} & \textcolor{red}{\xmark} & \textcolor{red}{\xmark} & \textcolor{red}{\xmark} & \textcolor{red}{\xmark} & \textcolor{green}{\cmark} \\
Training-free & \textcolor{green}{\cmark} & \textcolor{green}{\cmark} & \textcolor{green}{\cmark} & \textcolor{red}{\xmark} & \textcolor{red}{\xmark} & \textcolor{green}{\cmark} & \textcolor{green}{\cmark} & \textcolor{green}{\cmark} \\
Does not require source/target prompts & \textcolor{green}{\cmark} & \textcolor{green}{\cmark} & \textcolor{green}{\cmark} & \textcolor{green}{\cmark} & \textcolor{green}{\cmark} & \textcolor{red}{\xmark} & \textcolor{red}{\xmark} & \textcolor{red}{\xmark} \\
Does not require external segmentation model & \textcolor{green}{\cmark} & \textcolor{green}{\cmark} & \textcolor{red}{\xmark} & \textcolor{green}{\cmark} & \textcolor{green}{\cmark} & \textcolor{green}{\cmark} & \textcolor{green}{\cmark} & \textcolor{green}{\cmark} \\
\hline
\end{tabular}
}
\label{tab:baseline_comparison}
\vspace{-0.4cm}
\end{table*}

\section{Additional Analysis}
\label{sec:analysis}

\subsection{Motivation Validation} 
\begin{figure}[H]
  \centering
  \vspace{-0.1cm}
  \begin{subfigure}[b]{0.63\linewidth}
    \includegraphics[width=\linewidth]{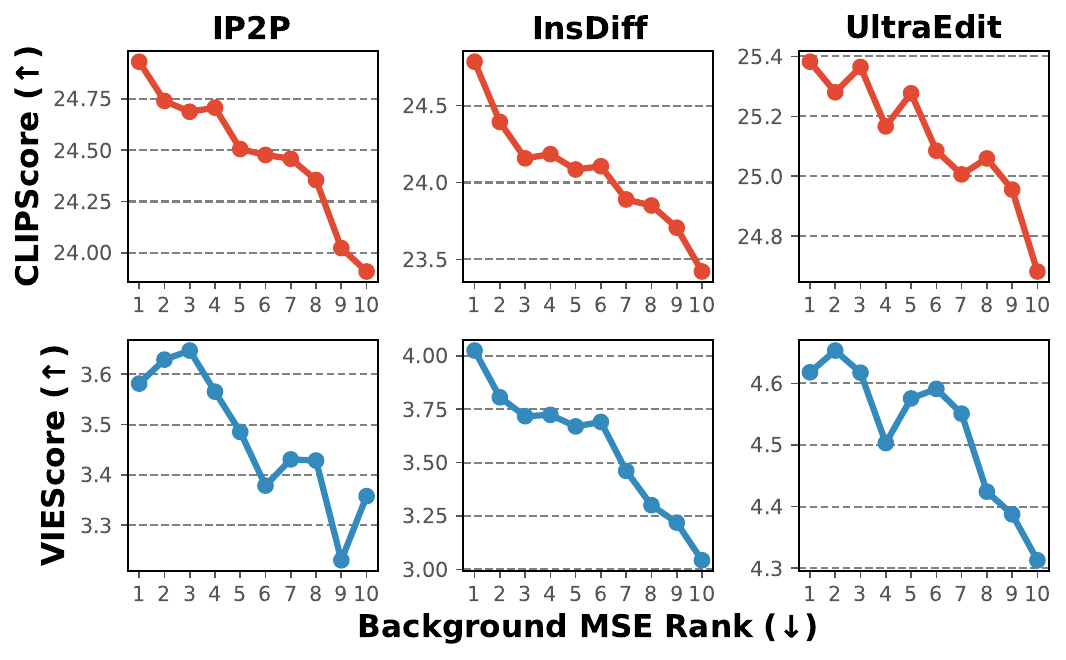}
    \vspace{-0.55cm}
    \caption{}
    \label{fig:motivation}
  \end{subfigure}
  \hfill
  \begin{subfigure}[b]{0.35\linewidth}
    \includegraphics[width=\linewidth]{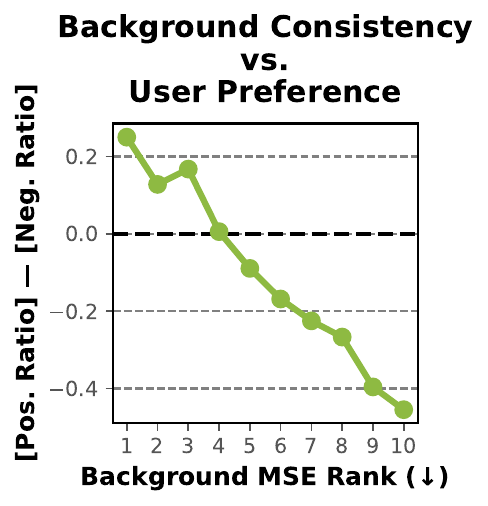}
    \vspace{-0.55cm}
    \caption{}
    \label{fig:user_study}
  \end{subfigure}
  \vspace{-0.15cm}
  \caption{(a) Higher background consistency (lower BG-MSE rank) correlates with higher CLIPScore and VIEScore across models. \textbf{(b) User Study:} User preference strongly correlates with background consistency (Pearson $r=0.534$), confirming perceptual alignment.
  } 
\end{figure}

We ranked each sample’s seed outputs by background MSE—treating the background as the preserved region—and found that lower background MSE strongly correlates with higher edit quality and better instruction following. \cref{fig:motivation} quantifies this relationship, while a complementary user study in \cref{fig:user_study} further corroborates it. In that study, 34 participants evaluated 52 high-variance, category-balanced PIE-Bench samples, each presenting ten outputs (seeds 1–10) produced by IP2P, InsDiff, or UltraEdit. Participants tagged each image as well- or poorly-edited, and we defined user preference as the difference between positive and negative responses for each rank. Inter-rater agreement was substantial (Krippendorff's $\alpha$=0.7535), underscoring the reliability of the findings. Moreover, \cref{fig:metric_w_mask} shows clear performance gains as background MSE decreases, confirming that background-consistent edits inherently yield superior results.

\subsection{Analysis of Timestep for Selection}

\begin{figure*}[t]
    \centering
    \includegraphics[width=\linewidth]{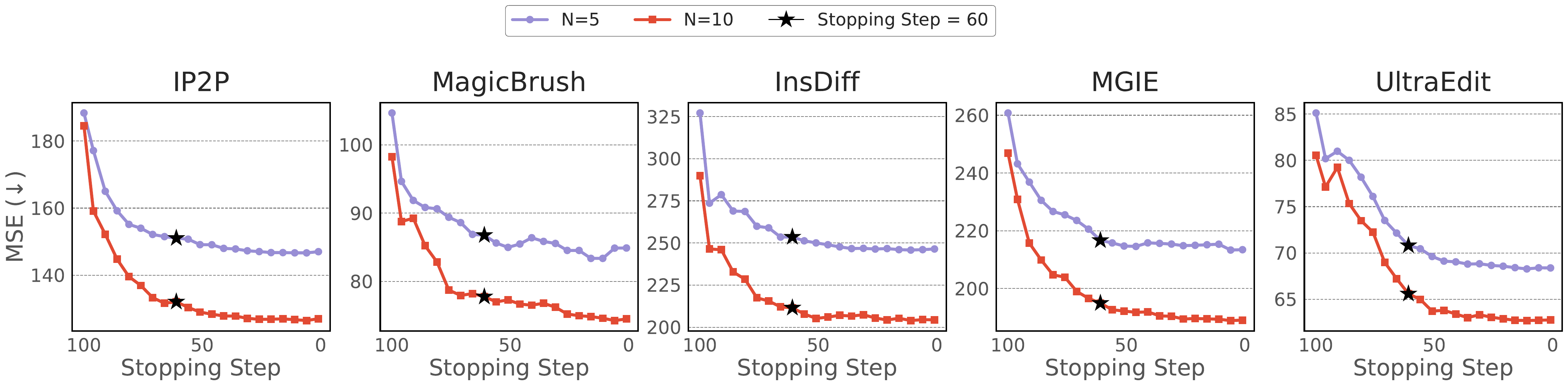}
    \caption{\textbf{ELECT performance variation with respect to stopping timestep ($t_\text{stop}$) with fixed number of seeds}.}
    \label{fig:stopping_steps_full}
\end{figure*}

We summarized our considerations regarding $t_\text{stop}$ in Section 5.4. Empirically, we observed that when $t_\text{stop}=60$, performance improvement began to converge across all models. In practice, stopping at this timestep resulted in balanced performance and efficiency gains. However, as shown in \cref{fig:stopping_steps_full}, for some models, $t_\text{stop}=60$ is not the optimal stopping step.

For instance, in the cases of IP2P and InsDiff, performance continues to converge sufficiently even at $t_\text{stop}=70$. By stopping at this point and performing selection, we can obtain output with fewer NFE while maintaining similar performance. We also identified a significant correlation between the convergence point of performance and the convergence point of changes in $S^\text{BIS}$, as shown in \cref{fig:result_grid}.

This phenomenon can be explained by the denoising process in image generation. In the early timesteps, images are heavily noisy, making it difficult to extract clean outputs that closely resemble the final result. However, beyond a certain point, the noise level decreases, and the model focuses on fine-grained details, leading to a stage where score variations become less significant.

Based on this observation, we argue that this specific point is where ranking the outputs produces minimal differences. Accordingly, we propose a criterion for determining a model- and sample-agnostic stopping step, which can be utilized for optimizing the selection process effectively.

Using a representative score \( S_t = \min_{i\in S} S^{\text{BIS}}(i,t) \) and its change \( \Delta S_t = |S_t - S_{t-1}| \), DDC stops denoising when the relative change \( \Delta S_t / \Delta S_{\max} \) falls below a threshold \( \tau \). With \( \tau = 0.1 \), UltraEdit converges at \( t_{\text{stop}} = 60 \), while other models converge near \( t_{\text{stop}} = 70 \), maintaining performance in fewer steps for some models (\cref{fig:result_grid}). In a 100-step process, heuristically setting $t_{\text{stop}}=60$ works broadly, though earlier stops (e.g., 70 or 80) suffice for some models without significant degradation.

\begin{figure}[H]
  \vspace{-0.2cm}
  \centering
  \includegraphics[width=0.7\linewidth]{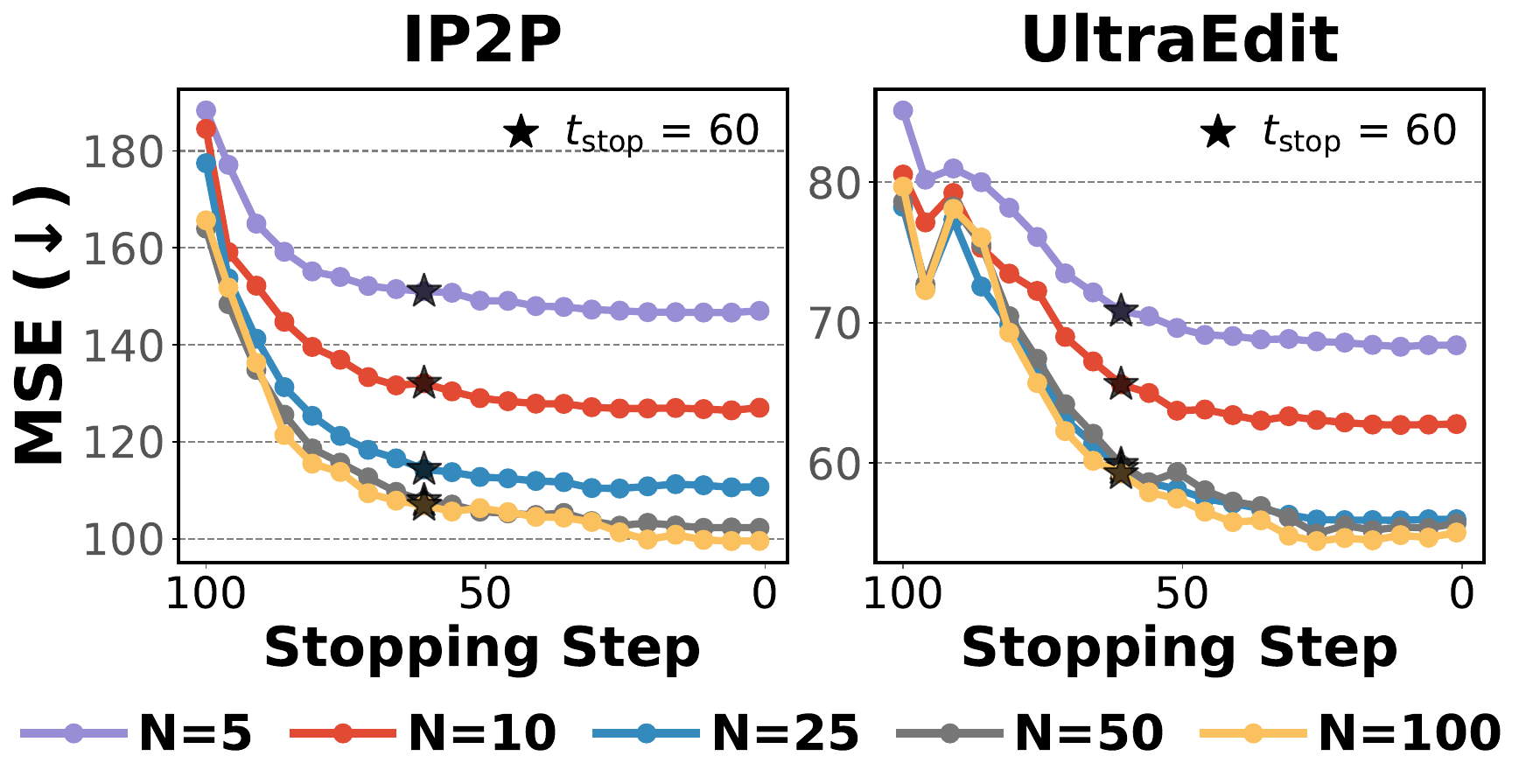}
  \vspace{-0.2cm}
   \caption{Performance trend in \cref{fig:stopping_steps} when N=25, 50, 100.}
   \label{fig:larger_N_fig7}
\end{figure}

We further examine how the benefits of \textsc{ELECT} scale with the number of candidate seeds~$N$. As shown in \cref{fig:larger_N_fig7}, the marginal performance improvements steadily taper off as $N$ grows, yet they remain consistently positive relative to the fixed-seed baseline. This figure extends the saturation trend observed in \cref{fig:stopping_steps} to a broader range of $N$ values, confirming that larger candidate pools yield diminishing—but still meaningful—returns.

\subsection{Analysis of Mask Extraction}

\begin{figure}[h]
    \centering
    \includegraphics[width=\linewidth]{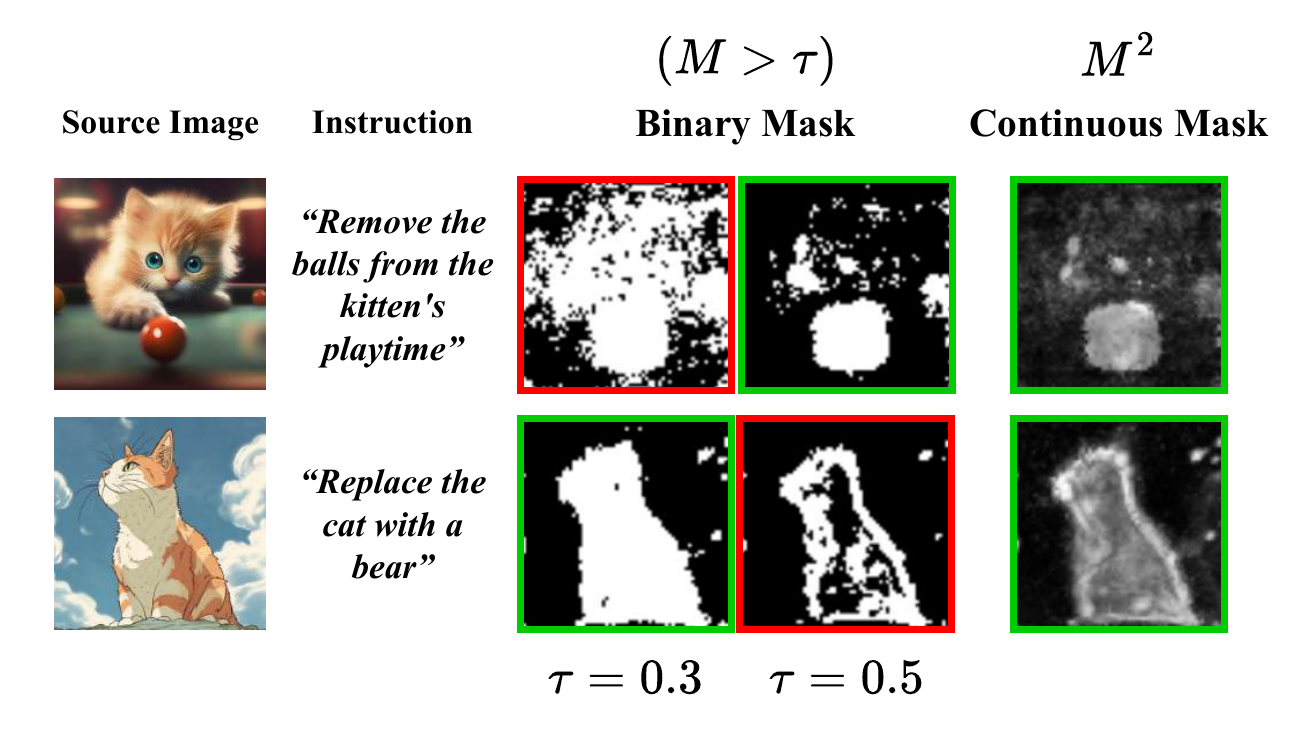}
    \caption{We further identified that the suitability of binary masks, derived from applying a threshold, varies significantly across samples. In contrast, the continuous mask consistently extracts stable regions of interest, as validated through our experiments.}
    \label{fig:mask_suppl}
\end{figure}

In prior work~\cite{Watch_Your_Steps_mirzaei2025watch}, relevance maps were extracted and subsequently binarized using a threshold before being utilized. However, we observed that the optimal threshold value varies across samples. Applying a fixed threshold for binarization often results in inaccurate mask extraction for certain samples, which in turn hinders the accurate computation of scores. Recognizing this limitation, we propose an approach that avoids hyperparameter tuning and instead leverages the continuous-valued mask directly to compute scores for regions outside the area of interest. As demonstrated in~\cref{fig:mask_suppl}, threshold-based methods exhibit a variety of failure cases depending on the chosen threshold. In contrast, our continuous mask assigns relatively higher real-valued scores to regions most relevant to editing. Consequently, when applying pixel-wise weighting, our method effectively penalizes background inconsistencies, offering a more robust solution.

To enhance this approach, we squared the mask values, which sharpens the distinction of regions outside the area of interest. This additional step amplifies the penalty on irrelevant areas, enabling a sample-robust application of the mask without the need for threshold adjustments.

\subsection{Analysis for Global Edits.}

\begin{figure}[H]
  \vspace{-0.2cm}
  \centering
  \includegraphics[width=\linewidth]{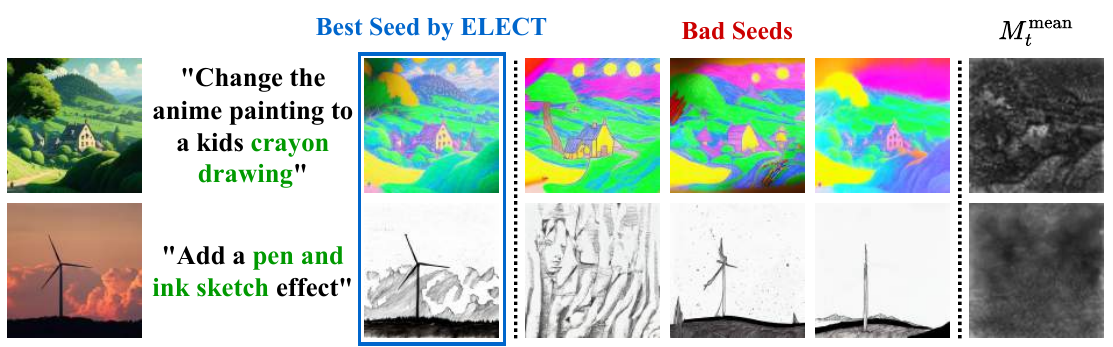}
  \vspace{-0.2cm}
   \caption{\textbf{Qualitative results of global editing with ELECT}.}
   \label{fig:global_editing}
\end{figure}

In global edits, ELECT selects seeds that better preserve the structural integrity of the source image while applying the intended style change. This benefit is evident in the mean relevance map \(M_t^{\text{mean}}\), which emphasizes broad, image-wide coherence rather than localized focus regions (see \cref{fig:global_editing,fig:overall_pipeline}). On \mbox{PIE-Bench}’s \textit{style transfer} task, \textsc{ELECT} (\(N\!=\!10\)) consistently outperforms fixed-seed baselines—reducing MSE and increasing SSIM for IP2P (\(\downarrow26\%\,/\,\uparrow3.4\text{ pt}\)), MagicBrush (\(\downarrow25\%\,/\,\uparrow6.6\text{ pt}\)), and UltraEdit (\(\downarrow19\%\,/\,\uparrow4.2\text{ pt}\)). It also improves semantic metrics, boosting CLIPScore by \(+0.01\text{–}0.83\) pt and VIEScore by \(+0.11\text{–}0.46\) pt across all models.

\subsection{Failure Cases.}

\begin{figure}[h]
  \centering
  \vspace{-0.2cm}
  \includegraphics[width=\linewidth]{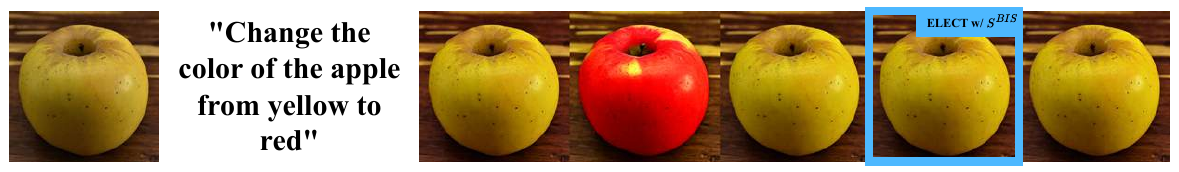}
  \vspace{-0.2cm}
   \caption{\textbf{A failure case of ELECT}.}
   \label{fig:over-optimization}
\end{figure}

Although \textsc{ELECT} can occasionally select edits that are overly mild—preserving too much of the background and dampening the intended change (see \cref{fig:over-optimization})—we observed that such instances were relatively uncommon in our experiments. This rarity likely stems from the strong modification bias of many instruction-guided image-editing models, which tends to push outputs toward more pronounced alterations, making excessive, unintended changes the more prevalent concern in practice.

\subsection{Another Signal for ELECT.}

\begin{figure}[h]
  \centering
  \vspace{-0.2cm}
  \includegraphics[width=\linewidth]{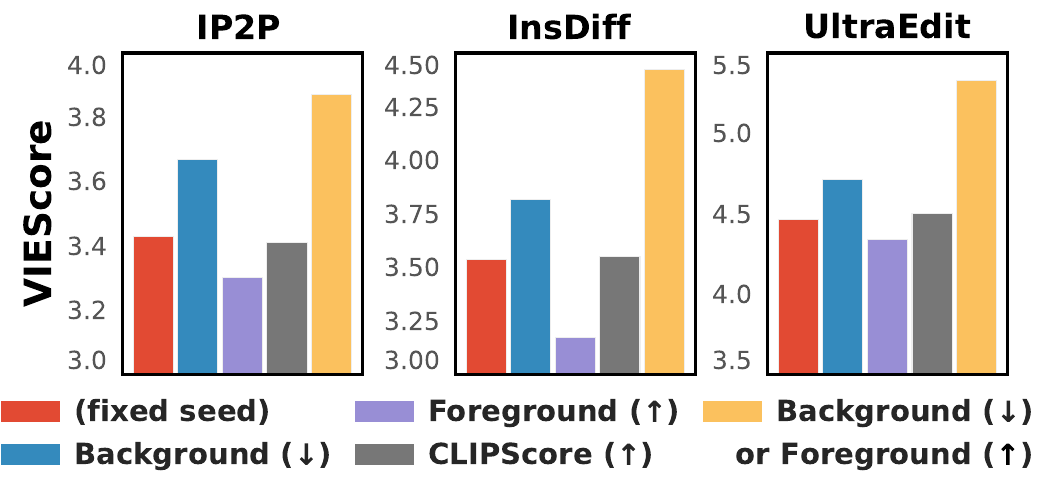}
  \vspace{-0.35cm}
   \caption{\textbf{Comparison with various signals for ELECT}.}
   \label{fig:another_signal}
\end{figure}

We also evaluated two alternative early-step cues for \textsc{ELECT}: foreground MSE (FG MSE) and CLIP-based text–image alignment. As illustrated in \cref{fig:another_signal}, both alternatives actually reduced performance—CLIP struggles with highly noisy early-timestep latents, and selecting the seed with the \emph{highest} FG MSE often drives excessive foreground changes, yielding over-edited images. A simple hybrid rule mitigates these issues: we choose between the seed with the lowest BG MSE and the one with the highest FG MSE, whichever produces the better preliminary score. This strategy (yellow bars) enhances robustness, rescuing failure cases where pure background consistency alone falters (see \cref{fig:over-optimization}). These observations highlight the opportunity to combine our pixel-wise background metric with complementary cues—such as foreground change or structural similarity—to support a more reliable, multi-aspect selection process. Realizing such a multi-objective framework will require deeper analysis, which we regard as a promising direction for future work.

\section{Extending Relevance Maps to Rectified Flow}
\label{app:rectified_flow}

Rectified Flow \cite{Rectified_Flow_liu2022flow} models such as Stable Diffusion 3 \cite{SD3_esser2024scaling} offer an alternative approach to modeling the noise-to-data transformation. The transformation is represented as an ordinary differential equation over a continuous time interval $t\in[0,1]$:
\begin{equation}
    dz_t=v(z_t,t)dt
\end{equation}
where $z_0\sim\pi_0$ is initialized from the source (noise) distribution and $z_1\sim\pi_1$ is generated at the end of the trajectory. The drift $v$ is fit to approximate the linear direction $z_1-z_0$:
\begin{equation}
    v_\theta(z_t,t)\simeq z_1-z_0
\end{equation}
Rectified flow models can also predict the denoised latent from timestep $t$ via 
\begin{equation}
    \hat{z}_0=z_t-v_\theta(z_t,t,I,C_T)\cdot t
\end{equation}
which corresponds to Tweedie's formula for diffusion models.

\begin{figure*}[tp]
    \centering
    \includegraphics[width=\textwidth]{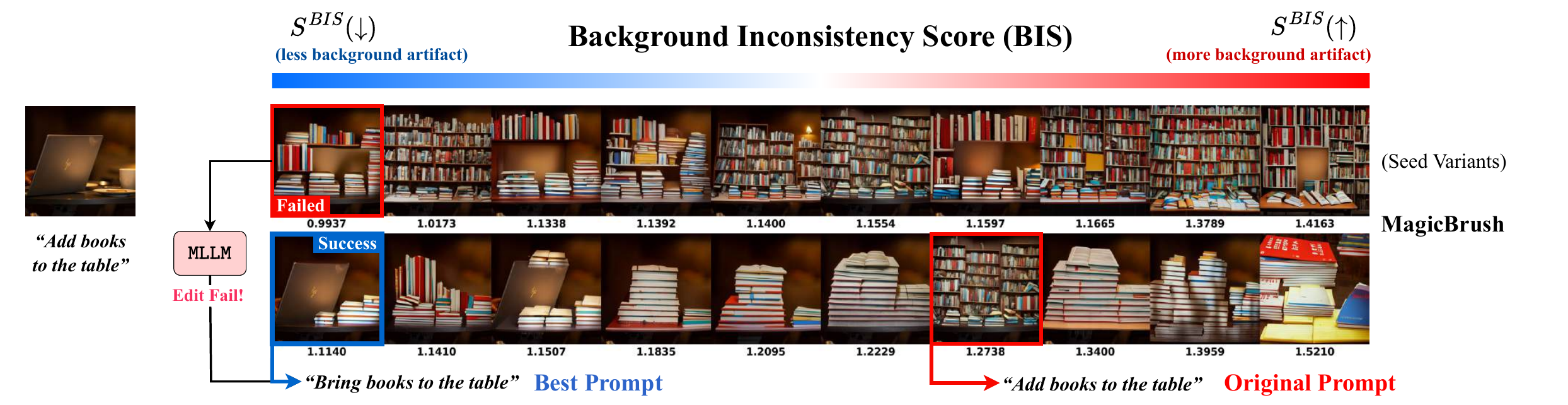}
    \caption{\textbf{ELECT extends to prompt selection by incorporating MLLMs, improving editing reliability when seed selection alone is insufficient.}}
    \label{fig:prompt_selection_example}
\end{figure*}

\section{ELECT for Instruction Prompt Selection}
\label{app:prompt_selection}

\begin{algorithm}[h]
\caption{$\texttt{ELECT}(\mathbb{S},t_\text{stop},\texttt{MLLM})=x^*$}
\label{alg:elect+mllm}
\begin{algorithmic}[1]
\Require Source image $I$, edit instruction $C_T$, candidate seed set $\mathbb{S}$, stopping timestep $t_{\text{stop}}$, instruction-guided denoiser $\epsilon_\theta$, VAE encoder $\mathcal{E}$ and decoder $\mathcal{D}$, MLLM $\mathcal{M}_\phi$
\Ensure Best edited image $x^*$
\State $x^0\gets \texttt{ELECT}(\mathbb{S},t_\text{stop})$ \Comment{\cref{alg:elect}}
\If {$\mathcal{M}_\phi(I,C_T,x^0,{\texttt{"evaluate $x^0$"}})>0$}
    \State \Return $x^*\gets x^0$ \Comment{Exit on edit success}
\EndIf
\State Sample a single initial noise $z_T \sim \mathcal{N}(0,I)$
\State $z_T^1=\cdots =z_T^N\gets z_T$
\State $\{C_i\}_{i=1}^N\gets \mathcal{M}_\phi (I,C_T,\texttt{{"generate $N$ prompts"}})$
\For{$t = T \to t_{\text{stop}}+1$} \Comment{Denoise until stopping time}
    \For{$i\gets 1,2,\dots,N$}
        \State $z_{t-1}^i \gets \texttt{Denoise}(z_t^i,t,I,C_i)$ 
    \EndFor
\EndFor
\For{$i \gets 1,2,\dots,N$}
    \State $S^\text{BIS}(i,t_\text{stop})\gets S^\text{BIS}(i,t_\text{stop}\mid [N],\epsilon_\theta,I,C_i)$
\EndFor
\State $i^* \gets \arg\min_{i \in [N]} S^\text{BIS}(i, t_{\text{stop}})$ \Comment{Select best prompt}
\For{$t = t_{\text{stop}} \to 1$} \Comment{Continue denoising $i^*$}
    \State $z^{i^*}_{t-1} \gets \texttt{Denoise}(z^{i^*}_{t},t,I,C_{i^*})$
\EndFor
\State \Return $x^* \gets \mathcal{D}(z^{i^*}_{0})$ \Comment{Final edited image}
\end{algorithmic}
\end{algorithm}

\paragraph{MLLM-Based Evaluation Metric.} To assess the success of image edits, we introduce an MLLM-based evaluation metric inspired by VIEScore \cite{Ku2023VIEScoreTE} and ImagenHub \cite{ku2024imagenhub}. While VIEScore provides a continuous score (0–10) for various aspects of an image, it lacks a definitive threshold for determining success. To address this, we adopt a discretized classification similar to ImagenHub, categorizing edits into three levels:
\begin{description}
    \item [1.0 (Success)] The edit fully satisfies the given instruction while maintaining background consistency.
    \item [0.5 (Partial Success)] The edit captures part of the instruction’s intent but introduces inconsistencies or artifacts.
    \item [0.0 (Failure)] The edit either does not follow the instruction or severely distorts the original image.
\end{description}
Following VIEScore's semantic consistency evaluation, we separately assesstwo key aspects:
\begin{enumerate}
    \item \textbf{Instruction Following}: Measures how well the edit aligns with the given prompt.
    \item \textbf{Background Consistency}: Ensures that unedited regions of the image remain unchanged.
\end{enumerate}
If either metric scores \textbf{0.0}, the edit is classified as a failure, triggering the prompt selection process. We provide the useful prompt used for MLLM evaluation:

\begin{Verbatim}[fontsize=\small, frame=single, breaklines=true]
"""
RULES:

Two images will be provided: The first being the original image and the second being an edited version of the first.
The objective is to evaluate how successfully the editing instruction has been executed in the second image. Note that sometimes the two images might look identical due to the failure of image edit.
To standardize the conduction of a rigorous human evaluation, we stipulate the criteria for each measurement as follows:  
Instruction Following (IF), score in range [0, 0.5, 1]
Background Consistency (BC), score in range [0, 0.5, 1]  

Instruction Following (IF) ensures that the generated image accurately follows the given editing instruction. In other words, the image has to be aligned with the requirements provided in user's inputs.
Background Consistency (BC) ensures that only the specified editing regions are modified, while unedited regions remain visually consistent with the original input image. This measures whether the image maintains fidelity in areas not targeted for editing.  

General Rules for Instruction Following (IF) scoring:  
IF=0: The scene in the edited image does not follow the editing instruction at all. IF=0.5: The scene in the edited image partially follows the editing instruction. IF=1: The scene in the edited image follows more than 75% of the editing instruction, aligning well with the intended changes. You agree that the overall idea is correct.  

General Rules for Background Consistency (BC) scoring:  
BC=0: Unedited regions are heavily altered, showing significant changes unrelated to the prompt or intended editing task. BC=0.5: Unedited regions are partly preserved, but some visible alterations or inconsistencies exist in areas that should remain unchanged. BC=1: Unedited regions are well-preserved, with no noticeable alterations or inconsistencies compared to the original input image.  

Scoring Criteria:  
Each metric (IF, BC) is independently scored, and the final evaluation is based on the aggregate results. High scores in all metrics indicate that the generated image successfully aligns with the prompt, maintains photorealism, and preserves the integrity of unedited regions.  

Return your evaluation in the following JSON format:  
{{
    "IF": <IF score>,
    "BC": <BC score>
}}
"""
\end{Verbatim}

\paragraph{Prompt Selection via MLLM.} For failed cases, we introduce an additional step where an MLLM generates alternative instruction prompts (\cref{fig:prompt_selection_example}). Given the input image and the original prompt, the MLLM is instructed to produce semantically equivalent but lexically varied instructions. To ensure diversity, we explicitly include constraints in the prompt, encouraging variations in wording, phrasing, and structure without altering the intended meaning.

This iterative process improves the likelihood of finding a prompt that falls within the model’s learned distribution, ultimately increasing the success rate of edits. The instruction generation prompt are provided below:

\begin{Verbatim}[fontsize=\small, frame=single, breaklines=true]
"""
You are an AI that generates editing instruction variants for text-guided image editing. Each variant should rephrase the editing instruction in a different way while strictly maintaining the original intent. Follow the given guidelines:

The input consists of:
1. A source image, which serves as the context for the editing instruction.
2. An editing instruction, describing the intended change to be made to the source image.

Your task is to create 10 diverse rephrasings of the editing instruction while preserving its original meaning.

### Guidelines:
1. The first variant should duplicate the given editing instruction exactly.
2. Subsequent variants should rephrase the instruction using different vocabulary, sentence structures, or expressions.
3. Ensure that all variants remain consistent with the source image and convey the same intent as the original instruction.
4. Avoid adding unnecessary complexity or details. Focus on concise and clear instructions.
5. Each instruction should be under 15 words and easy to understand.

### Input Example:
Source Image: (an image of a cat on a table)
Editing Instruction: "replace the cat with a dog"

### Output JSON Format:
{{
    "variants": [
        "replace the cat with a dog",
        "swap the cat for a dog",
        "make the cat a dog instead",
        ...
        "exchange the cat for a dog"
    ]
}}

### Note:
Ensure that all rephrasings align with the intent of the editing instruction while being consistent with the source image.

###Input:
Editing Instruction: {}
"""
\end{Verbatim}

\paragraph{Quantitative Results.} We evaluated PIE-bench data based on Background Consistency (BC) and Instruction Following (IF), categorizing each as 0, 0.5, or 1.0. Total number of data is 700 in PIE-bench. A case was considered a failure if either score was 0. We set the number of seeds to $N=10$ for ELECT and applied prompt selection only to the remaining failed cases after seed selection, with $N=10$ prompts for re-selection. As a result, the editing failure rate significantly decreased, successfully correcting approximately 40\% of previously failed baseline cases. (\cref{tab:mllm_evaluation}) Furthermore, we present the results of a comprehensive comparative evaluation of seed/prompt selection techniques across the whole metrics. (\cref{tab:mllm_quantitative})

\begin{table}[h]
\resizebox{\linewidth}{!}{
\begin{tabular}{c|c|c|c|c}
\cline{2-4}
\multicolumn{1}{l|}{} & \multicolumn{3}{c|}{Failure Ratio} & \multicolumn{1}{l}{} \\ \cline{2-5} 
\multicolumn{1}{l|}{} & \multicolumn{1}{c|}{Vanilla} & \multicolumn{1}{c|}{\begin{tabular}[c]{@{}c@{}}ELECT\\ (seed selection)\end{tabular}} & \multicolumn{1}{c|}{\begin{tabular}[c]{@{}c@{}}ELECT\\ (prompt selection)\end{tabular}} & \multicolumn{1}{c|}{\begin{tabular}[c]{@{}c@{}}Failure to Success\\ Ratio\end{tabular}} \\  \hline
\multicolumn{1}{|c|}{InstructPix2Pix} & 45.14\% & 40.00\% & \textbf{28.57\%} & \multicolumn{1}{c|}{36.71\%} \\ 
\multicolumn{1}{|c|}{MagicBrush} & 31.43\% & 26.71\% & \textbf{16.57\%} & \multicolumn{1}{c|}{47.27\%} \\ 
\multicolumn{1}{|c|}{InstructDiffusion} & 41.29\% & 34.29\% & \textbf{22.29\%} & \multicolumn{1}{c|}{46.02\%} \\
\multicolumn{1}{|c|}{MGIE} & 34.86\% & 33.00\% & \textbf{21.57\%} & \multicolumn{1}{c|}{38.11\%} \\ 
\multicolumn{1}{|c|}{UltraEdit} & 26.71\% & 23.43\% & \textbf{17.00\%} & \multicolumn{1}{c|}{36.36\%} \\ \hline
\end{tabular}
}
\caption{\textbf{Failure case analysis using the MLLM\cite{gpt4o} evaluator}.}
\label{tab:mllm_evaluation}
\end{table}

\begin{table*}
    \caption{\textbf{Comparison of prompt selection after seed selection and failed cases for ELECT seed selection.} The experiment was conducted with N=20 to ensure a fair comparison.Although selecting prompts after evaluating a larger number of seeds yields lower performance in terms of Background Consistency (BC), this does not necessarily translate to improved editing outcomes. As illustrated in~\cref{fig:main}, the performance tends to saturate, introducing a risk of over-optimization that may not lead to meaningfully better edits. In contrast, when prompt selection is performed after evaluating only 10 seeds and determining their failure, we observe improved performance in terms of Instruction Following. Notably, a significant increase in performance is evident when assessed using the VIEScore metric, which is known for its strong alignment with human judgment. This suggests that, for tasks that the model struggles to address under the initial prompt conditions, introducing an alternative signal enables a broader and more effective search for outputs closer to success. }

\resizebox{\textwidth}{!}{
    \begin{tabular}{l|c|cccc|c|ccc}
        \toprule
        \multirow{2}{*}{Model} & \multirow{2}{*}{\makecell{ Seed Selection \\ Method}} & \multicolumn{4}{c|}{BC} & \multicolumn{1}{c|}{IF} & \multicolumn{3}{c|}{VIEScore (Semantic Consistency) ($\uparrow$)} \\
        \cmidrule{3-10}
        & & $\text{MSE}_{\times10^4}$ ($\downarrow$) & $\text{LPIPS}_{\times10^3}$ ($\downarrow$) & PSNR ($\uparrow$) & $\text{SSIM}_{\times10^2}$ ($\uparrow$) & CLIP-T ($\uparrow$) & BC & IF & min(BC, IF)  \\
        \midrule 
\multirow{4}{*}{IP2P} & Vanilla & 248.49 & 162.41 & 20.73 & 75.98 & 24.38 & 6.02 & 4.15 & 3.43 \\
 & ELECT (seed $N=10$) & 128.80 & 104.25 & 23.28 & 80.86 & 24.93 & 6.80 & 4.27 & \underline{3.68} \\
 & ELECT (seed $N=20$) & \textbf{115.97} & \textbf{98.27} & \textbf{23.62} & \textbf{81.41} & \underline{24.95} & \textbf{6.97} & \underline{4.33} & 3.60 \\
 & ELECT (seed to prompt $N=20$) & \underline{127.18} & \underline{100.91} & \underline{23.48} & \underline{81.18} & \textbf{25.05} & \underline{6.85} & \textbf{4.65} & \textbf{3.92} \\
 \midrule
\multirow{4}{*}{MagicBrush} & Vanilla & 139.18 & 77.22 & 24.83 & 82.84 & 24.63 & 5.89 & 4.70 & 3.99 \\
 & ELECT (seed $N=10$) & \underline{75.75} & 59.57 & \underline{26.12} & 84.63 & 24.98 & 6.27 & 4.90 & 4.25 \\
 & ELECT (seed $N=20$) & \textbf{72.15} & \textbf{57.50} & \textbf{26.28} & \textbf{84.86} & \underline{25.03} & \underline{6.33} & \underline{4.99} & \underline{4.33} \\
 & ELECT (seed to prompt $N=20$) & 78.33 & \underline{58.63} & \underline{26.12} & \underline{84.68} & \textbf{25.15} & \textbf{6.55} & \textbf{5.30} & \textbf{4.58} \\
 \midrule
\multirow{4}{*}{InsDiff} & Vanilla & 372.46 & 154.04 & 20.25 & 75.53 & 24.09 & 5.42 & 4.18 & 3.53 \\
 & ELECT (seed $N=10$) & \underline{179.64} & \underline{103.91} & \underline{22.89} & \underline{80.09} & 24.71 & \underline{5.87} & 4.54 & 3.82 \\
 & ELECT (seed $N=20$) & \textbf{165.79} & \textbf{103.05} & \textbf{23.03} & \textbf{80.23} & \underline{24.87} & \underline{5.87} & \underline{4.62} & \underline{3.86} \\
 & ELECT (seed to prompt $N=20$) & 191.25 & 103.92 & 22.78 & 80.06 & \textbf{24.97} & \textbf{6.16} & \textbf{5.05} & \textbf{4.28} \\
 \midrule
\multirow{4}{*}{MGIE} & Vanilla & 341.42 & 145.51 & 21.16 & 77.31 & 24.44 & 5.64 & 4.41 & 3.68 \\
 & ELECT (seed $N=10$) & 187.40 & 103.61 & 23.54 & 81.27 & 24.68 & 6.27 & \underline{4.55} & \underline{3.93} \\
 & ELECT (seed $N=20$) & \underline{176.79} & \underline{98.24} & \underline{23.83} & \underline{81.73} & \underline{24.81} & \underline{6.30} & 4.52 & 3.91 \\
 & ELECT (seed to prompt $N=20$) & \textbf{137.01} & \textbf{88.40} & \textbf{24.22} & \textbf{82.59} & \textbf{25.10} & \textbf{6.55} & \textbf{4.88} & \textbf{4.21} \\
 \midrule
\multirow{4}{*}{UltraEdit} & Vanilla & 87.54 & 115.37 & 22.93 & 79.86 & 25.20 & 5.89 & 5.50 & 4.47 \\
 & ELECT (seed $N=10$) & \underline{64.20} & \underline{93.15} & \underline{24.46} & \underline{83.56} & \underline{25.37} & \underline{6.37} & \underline{5.63} & 4.71 \\
 & ELECT (seed $N=20$) & \textbf{60.28} & \textbf{89.53} & \textbf{24.76} & \textbf{84.07} & \textbf{25.51} & \textbf{6.47} & 5.62 & \underline{4.77} \\
 & ELECT (seed to prompt $N=20$) & 70.17 & 99.18 & 23.90 & 82.54 & 25.26 & 6.24 & \textbf{5.95} & \textbf{4.90} \\
 \bottomrule
    \end{tabular}%
    }
    \label{tab:mllm_quantitative}
\end{table*}

\begin{figure*}[bp]
    \centering
    \includegraphics[width=\textwidth]{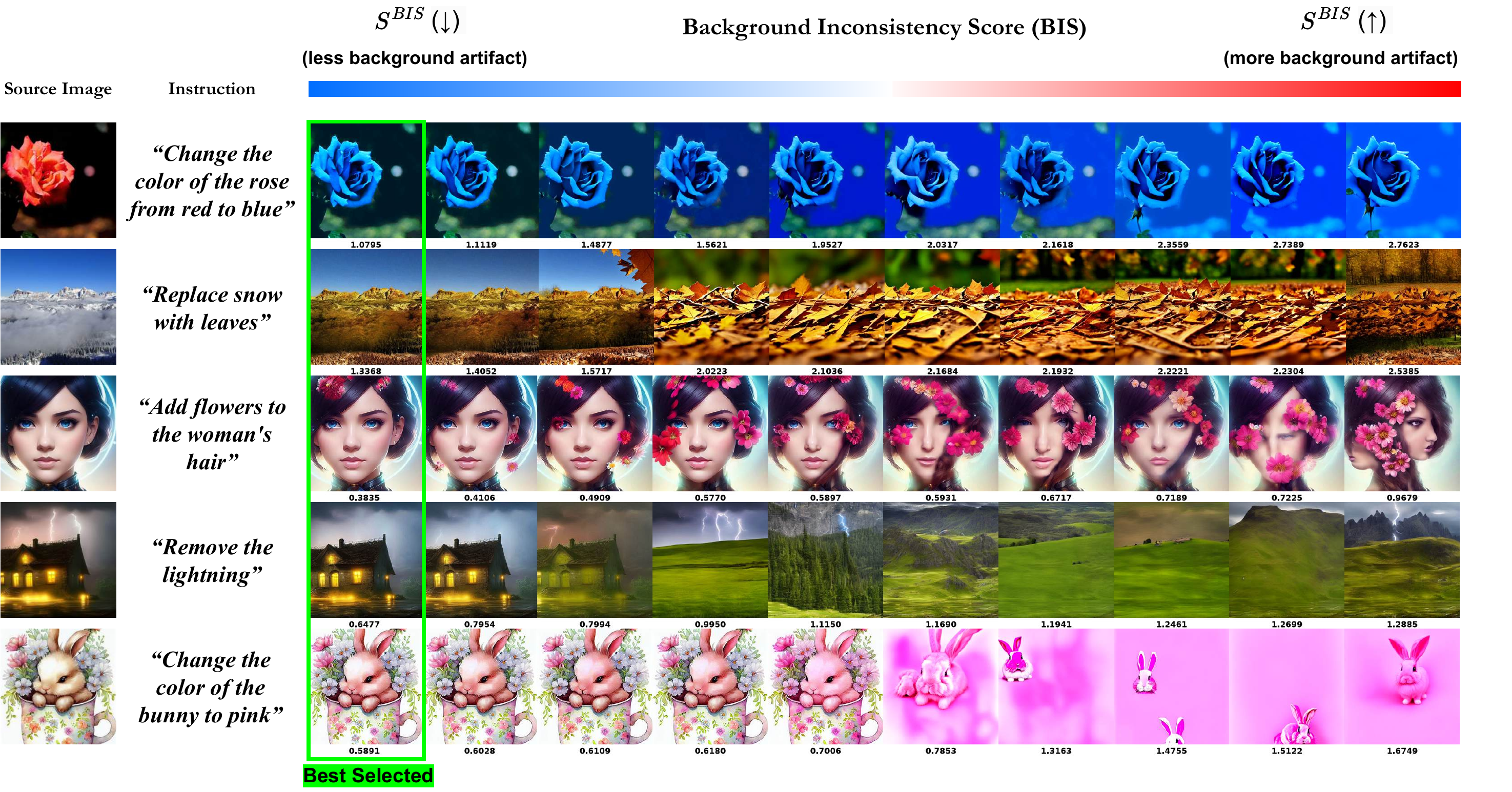}
    \caption{\textbf{Qualitative Result for Seed Selection} (dataset: PIE-bench \cite{PnP_Inversion_ju2023direct}, model: \textbf{InstructPix2Pix \cite{IP2P_brooks2023instructpix2pix})}.}
    \label{fig:qualitative_pie_ip2p}
\end{figure*}

\section{Additional qualitative results}
\label{app:additional_qualitative_results}
\begin{figure*}[h]
    \centering
    \includegraphics[width=\textwidth]{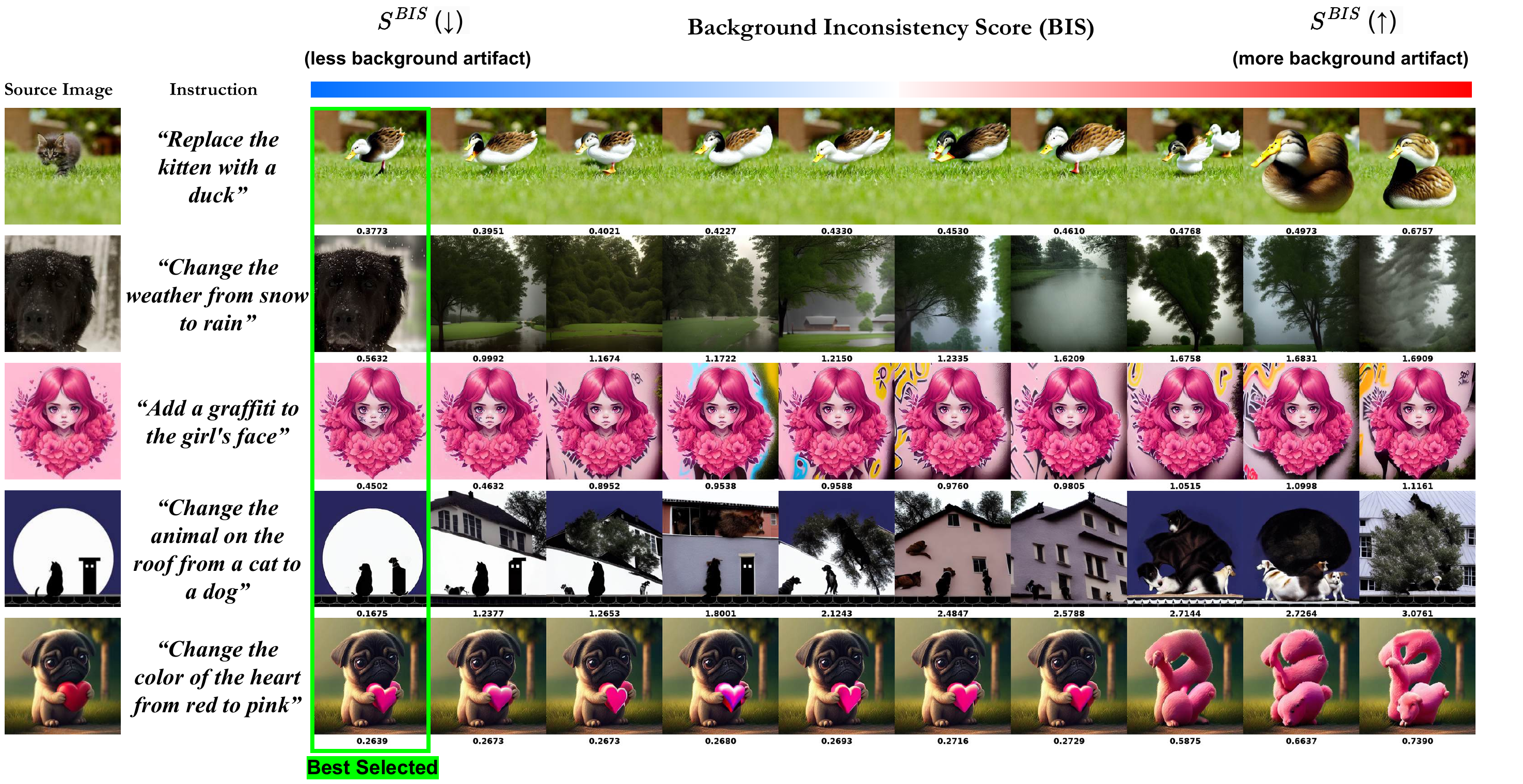}
    \caption{\textbf{Qualitative Result for Seed Selection }(dataset: PIE-bench \cite{PnP_Inversion_ju2023direct}, model:\textbf{ MagicBrush \cite{MagicBrush_NEURIPS2023_64008fa3})}.}
    \label{fig:qualitative_pie_mb}
\end{figure*}

\begin{figure*}[h]
    \centering
    \includegraphics[width=\textwidth]{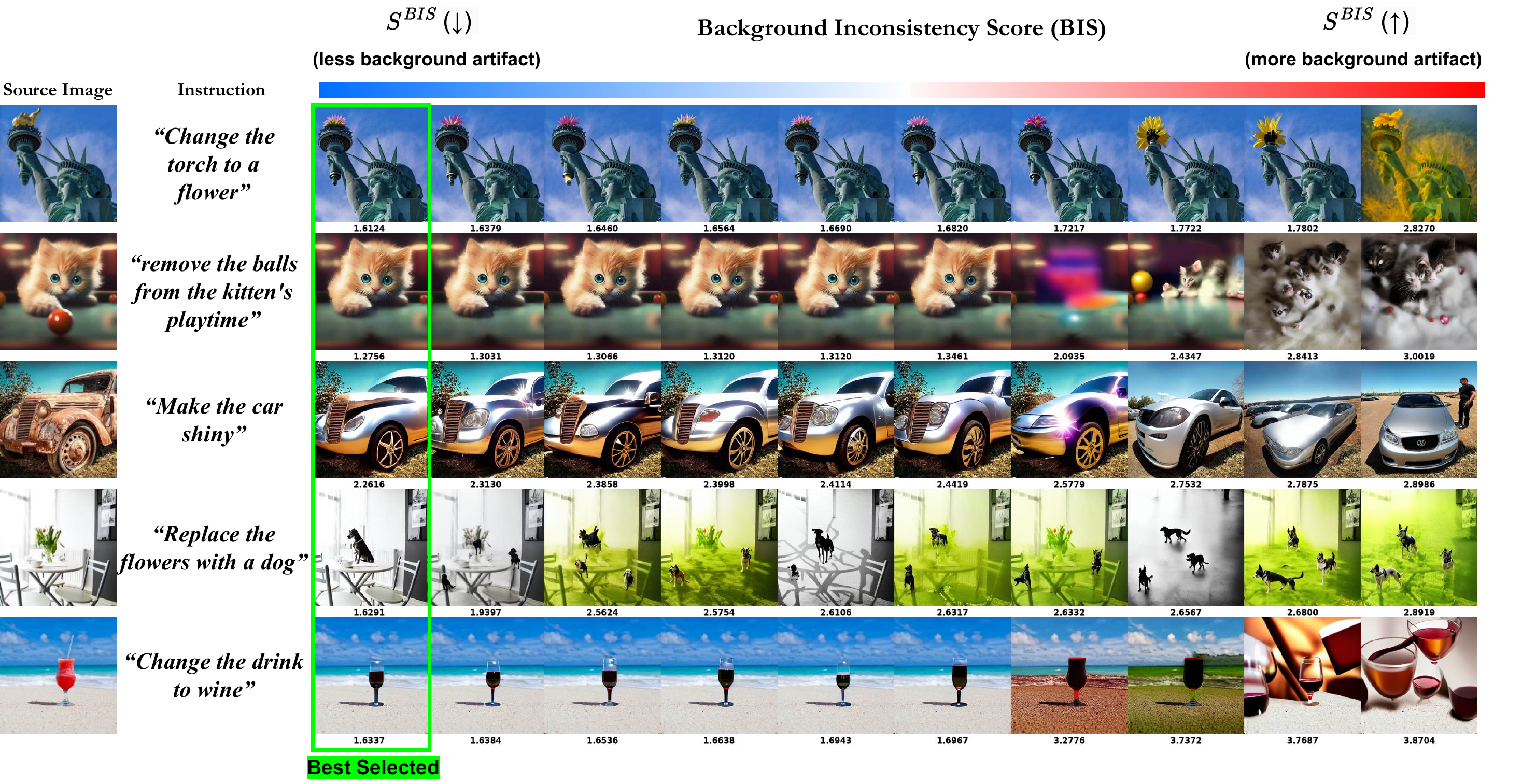}
    \caption{\textbf{Qualitative Result for Seed Selection }(dataset: PIE-bench \cite{PnP_Inversion_ju2023direct}, model:\textbf{ InstructDiffusion \cite{InstructDiff_Geng23instructdiff})}.}
    \label{fig:qualitative_pie_insdiff}
\end{figure*}

\begin{figure*}[h]
    \centering
    \includegraphics[width=\textwidth]{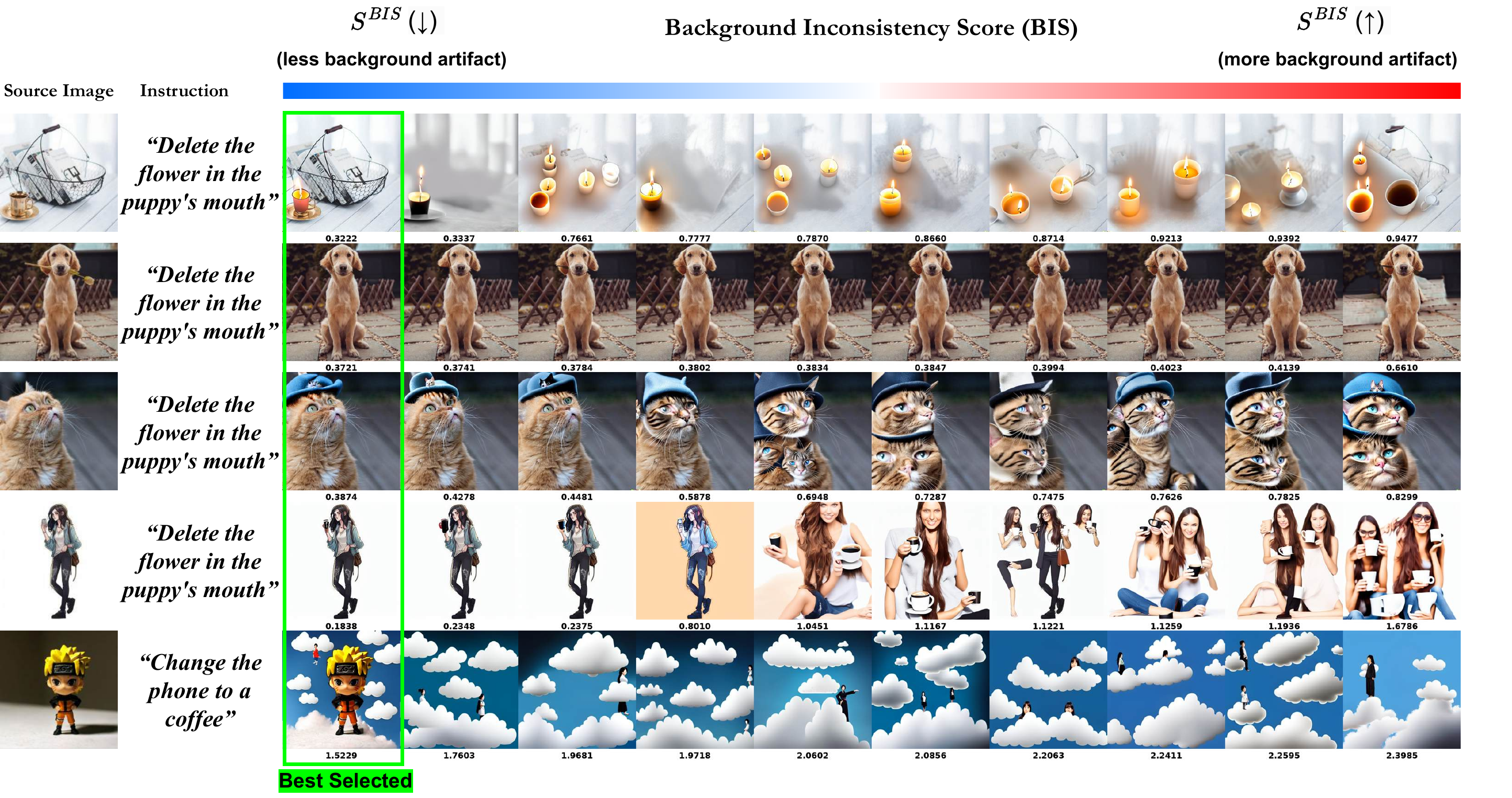}
    \caption{\textbf{Qualitative Result for Seed Selection }(dataset: PIE-bench \cite{PnP_Inversion_ju2023direct}, model:\textbf{ MGIE \cite{MGIE_fu2024guiding})}.}
    \label{fig:qualitative_pie_mgie}
\end{figure*}

\begin{figure*}[h]
    \centering
    \includegraphics[width=\textwidth]{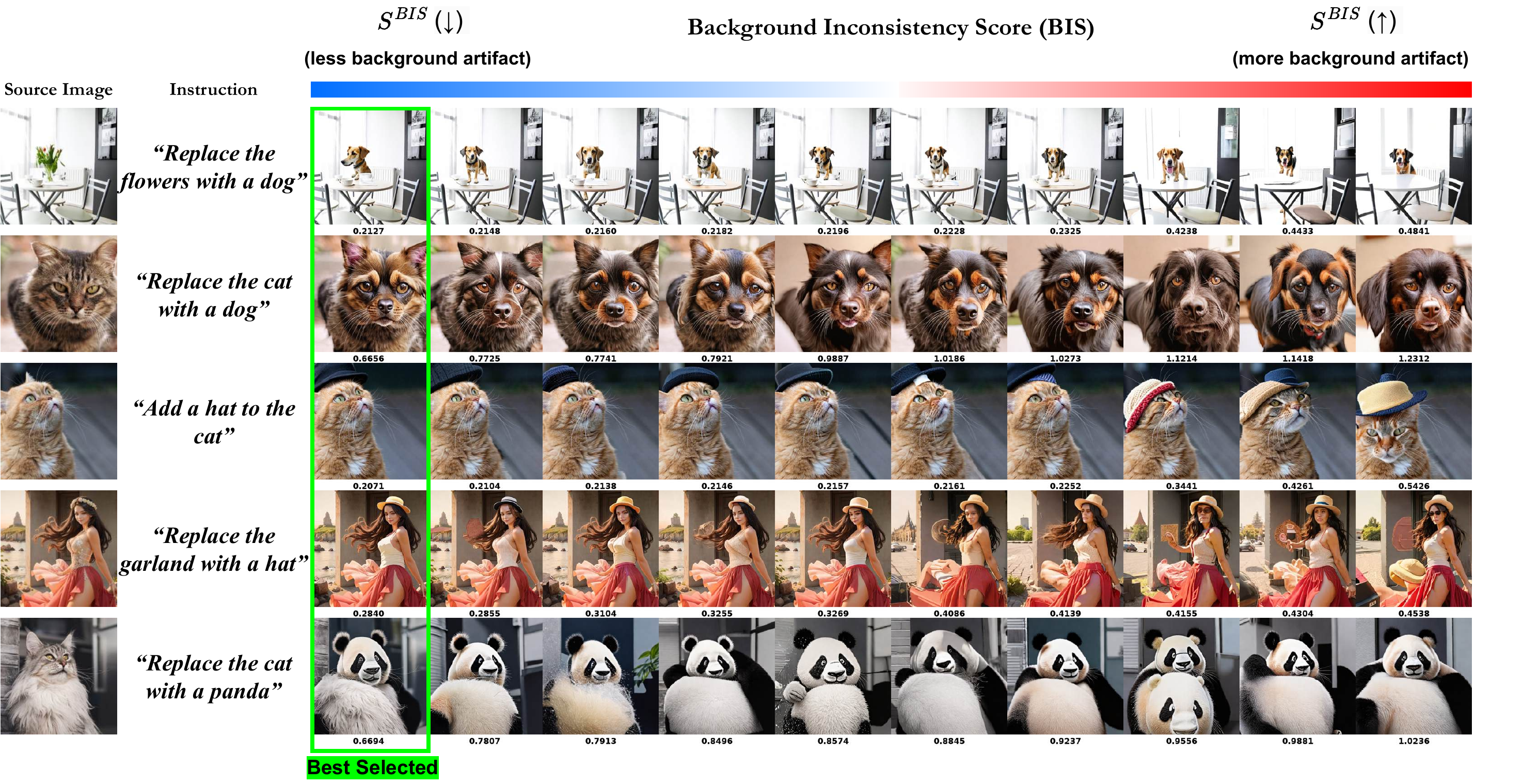}
    \caption{\textbf{Qualitative Result for Seed Selection }(dataset: PIE-bench \cite{PnP_Inversion_ju2023direct}, model:\textbf{ UltraEdit \cite{UltraEdit_zhao2024ultraedit})}.}
    \label{fig:qualitative_pie_ue}
\end{figure*}

\begin{figure*}[h]
    \centering
    \includegraphics[width=\textwidth]{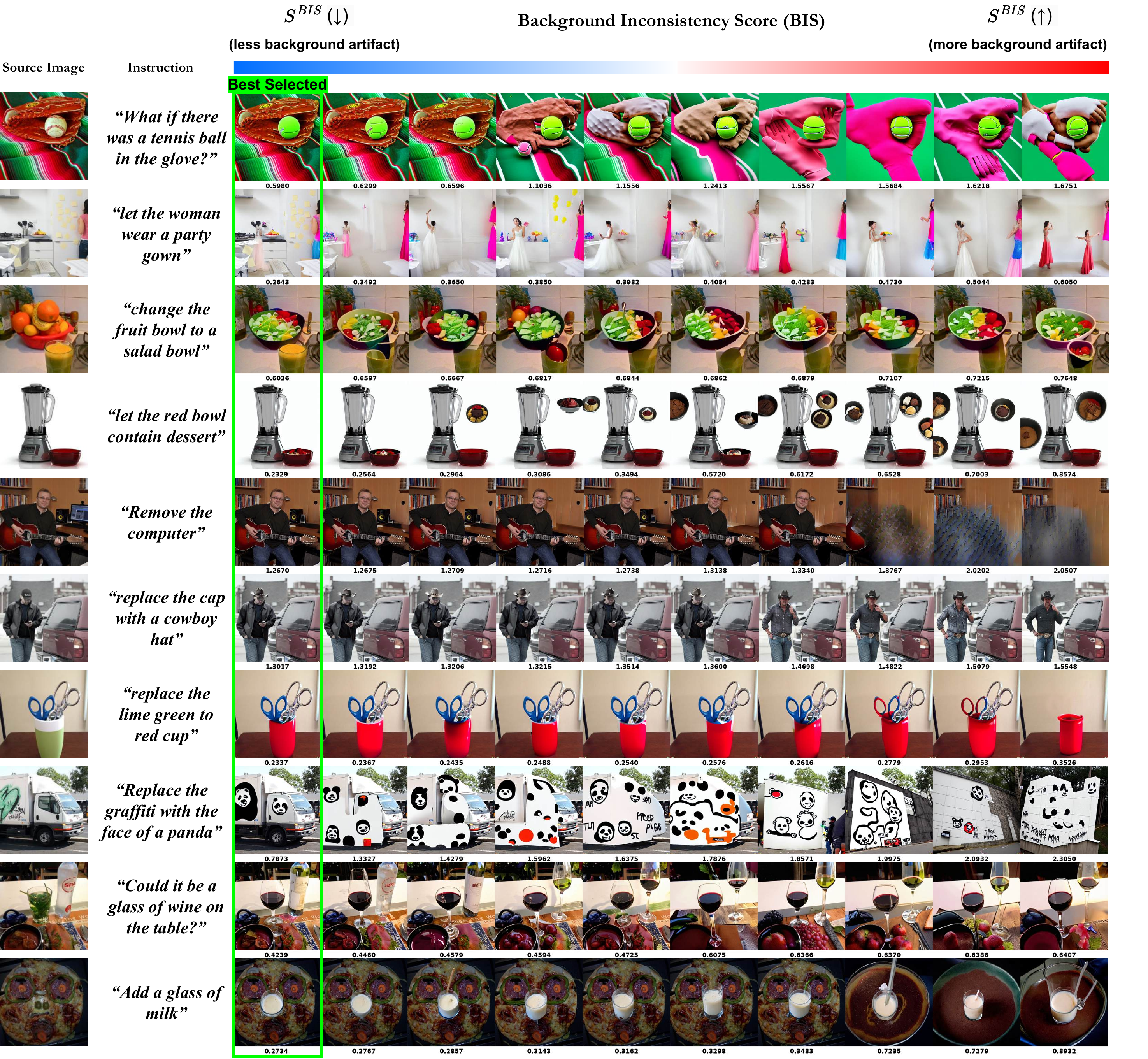}
    \caption{\textbf{Qualitative Result for Seed Selection (dataset: MagicBrush \cite{MagicBrush_NEURIPS2023_64008fa3})}. From top to bottom, each model’s results — InstructPix2Pix \cite{IP2P_brooks2023instructpix2pix}, MagicBrush \cite{MagicBrush_NEURIPS2023_64008fa3}, InstructDiffusion \cite{InstructDiff_Geng23instructdiff}, MGIE \cite{MGIE_fu2024guiding}, and UltraEdit \cite{UltraEdit_zhao2024ultraedit} — are displayed in order, with two rows per model.}
    \label{fig:qualitative_mb}
\end{figure*}

\begin{figure*}[h]
    \centering
    \includegraphics[width=\textwidth]{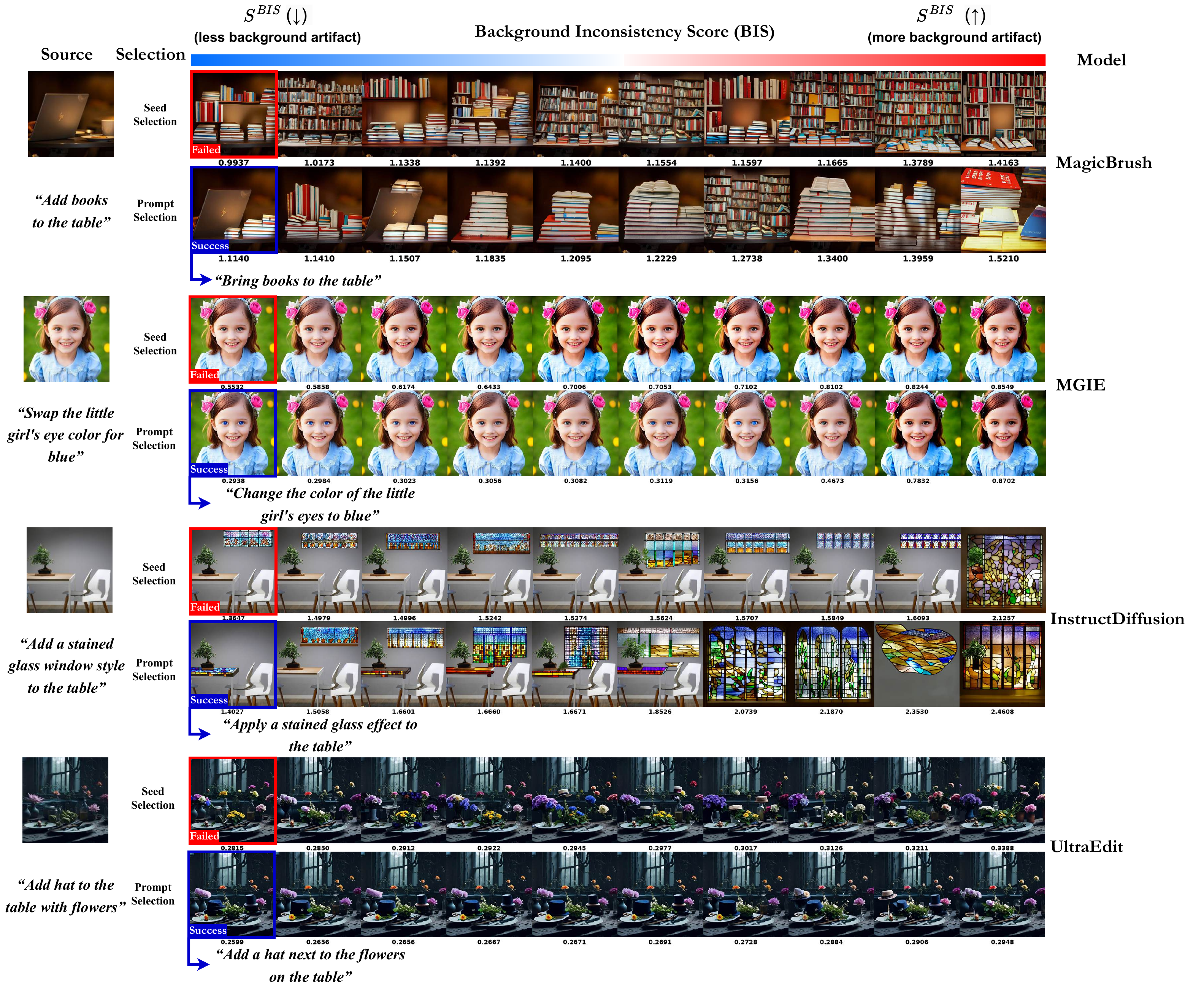}
    \caption{\textbf{Qualitative Result for Prompt Selection (dataset: PIE-bench \cite{PnP_Inversion_ju2023direct}).} MLLM-generated instruction variants refine failed edits to enhance overall editing outcomes.}
    \label{fig:qualitative_prompt_selection}
\end{figure*}

We provide various qualitative results for PIE-bench\cite{PnP_Inversion_ju2023direct} (\cref{fig:qualitative_pie_ip2p}, \cref{fig:qualitative_pie_mb}, \cref{fig:qualitative_pie_insdiff}, \cref{fig:qualitative_pie_mgie}, \cref{fig:qualitative_pie_ue}) and MagicBrush\cite{MagicBrush_NEURIPS2023_64008fa3} (\cref{fig:qualitative_mb}). Starting from the next image, the selected candidates using ELECT ($N=10$) are placed on the far left, and the sorted qualitative results, where the score increases (background inconsistency rises) towards the right, are shown. In addition, \cref{fig:qualitative_prompt_selection} illustrates cases where initial seed selection ($N=10$) failed but were successfully handled by prompt selection ($N=10$). In all qualitative results, the scores shown below each image correspond to $S^{BIS}$.

\end{document}